\documentclass{article}
\usepackage[dvipsnames]{xcolor}
\usepackage{microtype}
\usepackage{graphicx}
\usepackage{subfigure}
\usepackage{booktabs}
\usepackage{hyperref}

\usepackage[accepted]{icml2025}
\usepackage{amsmath}
\usepackage{amssymb}
\usepackage{mathtools}
\usepackage{amsthm}
\usepackage[capitalize,noabbrev]{cleveref}

\theoremstyle{plain}

\theoremstyle{definition}

\theoremstyle{remark}

\usepackage[textsize=tiny]{todonotes}

\usepackage{times}
\usepackage{latexsym}
\usepackage[T2A,T1]{fontenc}
\usepackage[utf8]{inputenc}
\usepackage{inconsolata}
\usepackage{xspace}
\usepackage{float} 
\usepackage{multirow}
\usepackage{algorithm}
\usepackage{algorithmic}
\usepackage{tabularx}
\usepackage[compatibility=false]{caption}
\DeclareUnicodeCharacter{2581}{\textbf{\_}}

\DeclareMathOperator*{\argmax}{argmax}
\DeclareMathOperator*{\Tok}{Tok}

\newcommand{\tokenizer}[2]{#1\textsubscript{#2}}
\newcommand{\lang}[2]{#1 $\rightarrow$ #2\xspace}

\newcommand{\spsrc}{\tokenizer{SP}{src}\xspace}
\newcommand{\sptgt}{\tokenizer{SP}{tgt}\xspace}
\newcommand{\sptrg}{\sptgt}
\newcommand{\psp}{\tokenizer{PairedSP}{}\xspace}
\newcommand{\pspem}{\tokenizer{PairedSP}{EM}\xspace}
\newcommand{\usp}{\tokenizer{PairedSP}{M}\xspace}

\newcommand{\vtgt}{V_\text{tgt}}
\newcommand{\vtrg}{\vtgt} % consistency
\newcommand{\vsrc}{V_\text{src}}

\newcommand{\gvnote}[1]{}
\newcommand{\gv}[1]{}

\newcommand{\jl}[1]{}
\newcommand{\MP}[1]{}

\usepackage{pgf}
\usepackage{colortbl}

\newcolumntype{C}{>{\centering\arraybackslash}X} 

\usepackage[ukrainian, english]{babel}
\usepackage{enumitem}

\newcommand{\CALL}[2]{\textsc{#1}(#2)}

\newcommand{\RETURN}{\STATE \textbf{return}\xspace}

\icmltitlerunning{Conditional Unigram Tokenization with Parallel Data}

\begin{document}

\twocolumn[
    \icmltitle{Conditional Unigram Tokenization with Parallel Data}
    \icmlsetsymbol{equal}{*}

    \begin{icmlauthorlist}
    \icmlauthor{Gianluca Vico}{ufal}
    \icmlauthor{Jindřich Libovický}{ufal}
    \end{icmlauthorlist}
    
    \icmlaffiliation{ufal}{Institute of Formal and Applied Linguistics, Faculty of Mathematics and Physics, Charles University, Prague, Czech Republic}
    
    \icmlcorrespondingauthor{Gianluca Vico}{vico [at] ufal.muff.cuni.cz}
    \icmlcorrespondingauthor{Jindřich Libovický}{libovicky [at] ufal.muff.cuni.cz}
    
    \icmlkeywords{Natural Language Processing, Tokenization, Alignment}
    
    \vskip 0.3in
]

\printAffiliationsAndNotice{}

\begin{abstract}
We introduce conditional unigram tokenization, a novel approach that extends unigram tokenization by conditioning target token probabilities on source-language tokens from parallel data.
Given a fixed source tokenizer, our method learns a target tokenizer that maximizes cross-lingual semantic alignment.
We evaluate our tokenizer on four language pairs across different families and resource levels, examining intrinsic properties and downstream performance on machine translation and language modeling.
While our conditional tokenizer maintains comparable statistical properties to standard unigram tokenizers, results are mixed: we observe no improvements in machine translation quality, but find consistent perplexity reductions in language modeling.
We hypothesize that quadratic scaling of conditional probability estimation with respect to the vocabulary size creates a data efficiency bottleneck.
Our findings suggest that alternative parameterizations may be necessary for practical cross-lingual tokenization.
\end{abstract}

\section{Introduction}

Tokenization serves as the foundation of most natural language processing pipelines, directly influencing model performance across tasks. While traditional tokenization approaches \citep{sennrich-etal-2016-neural,kudo-2018-subword} focus primarily on token frequency in monolingual contexts, their effectiveness in multilingual scenarios depends critically on achieving both literal \citep{pires-etal-2019-multilingual,limisiewicz-etal-2023-tokenization} and semantic \citep{hammerl-etal-2025-beyond} overlap between languages. Improving the semantic overlap of tokenizers in different languages might be beneficial, particularly for low-resource languages that suffer from low performance caused, among others, by overtokenization \citep{ahia-etal-2023-languages}. Therefore, these languages might benefit from cross-lingual alignability.

In this paper, we introduce a novel approach to cross-lingual tokenization that attempts to directly address this challenge in a probabilistic model. Given an existing tokenizer in a source language, we develop a target language tokenizer that maximizes semantic alignment between the two languages. Our approach extends the unigram tokenization framework \citep{kudo-2018-subword} by replacing unconditional unigram probabilities with conditional probabilities based on source-language tokens.

Specifically, we formulate tokenization as maximizing the unigram probability of target tokens conditioned on aligned source tokens from parallel data. It is a straightforward generalization of standard unigram tokenization, with the key difference that it explicitly models cross-lingual token alignability during the tokenizer training process. Similarly to the unigram model, this is also used for vocabulary learning.

We evaluate our approach on four language pairs across eight translation directions, analyzing both intrinsic tokenization properties and downstream task performance. Our results present a mixed picture: while the intrinsic evaluation shows that our conditional tokenizer maintains statistical properties comparable to standard unigram tokenizers, we do not observe consistent improvements in machine translation quality. However, we do find notable perplexity reductions in language modeling tasks, suggesting potential benefits for specific applications.

The remainder of this paper is organized as follows: Section~\ref{sec:method} details our conditional unigram tokenization approach. Section~\ref{sec:experiments} and ~\ref{sec:results} present experimental results across multiple language pairs and tasks.
Finally, Section~\ref{sec:conclusion} discusses implications and directions for future research.
The source code for replicating our experiments is openly available on GitHub (\url{https://github.com/GianlucaVico/Conditional-Unigram-Tokenization}).
% The source code for replicating our experiments is openly available on GitHub\footnote{Repository: \url{https://github.com/GianlucaVico/Conditional-Unigram-Tokenization}.}.

\section{Related Work}

\paragraph{Subword Tokenization.} The most frequently used subword tokenizers in NLP are BPE \cite{sennrich-etal-2016-neural} and Unigram \cite{kudo-2018-subword}. These approaches address out-of-vocabulary (OOV) words while maintaining a fixed vocabulary size and ensuring tokens have comparable frequencies for proper embedding training. These methods typically represent common words as single tokens, while rare words
% \MP{The following two (low-res-lang words and non-Latin) are rather examples of rare words, so I would put them e.g. in parenthesis and prefixed with "including". In other words, \textbf{frequent} words will \textbf{not} be fragmented, even if they are from low-res langs or non-Latin scripts.}
(including words from low-resource languages, or those in non-Latin scripts) get fragmented into multiple tokens or individual bytes \citep{petrov_language_2023,ahia-etal-2023-languages}. Notable alternative approaches include VOLT \cite{xu-etal-2021-vocabulary}, which employs optimal transport for vocabulary construction, or tokenization inference methods, such as PathPiece \cite{schmidt_tokenization_2024}, which generates the shortest possible token sequence for a given vocabulary, or Legros \citep{libovicky-helcl-2024-lexically} that finds the most semantically plausible tokenization for a given vocabulary.

\paragraph{Cross-lingual Token Alignment.} Previous studies \citep{minixhofer_wechsel_2022,remy_tik--tok_2023,remy_trans-tokenization_2024} showed that token semantic similarity across languages is important for effective cross-lingual transfer. This similarity can be derived from bilingual dictionaries \cite{minixhofer_wechsel_2022} or through automated techniques \cite{remy_trans-tokenization_2024}, such as Fast Align \cite{dyer-etal-2013-simple}. \citet{hammerl-etal-2025-beyond} establish that token alignment between parallel sentences correlates with performance on multiple downstream tasks and introduces metrics for measuring such alignment across different tokenizers using a statistical model for word alignment.

\paragraph{Joint Tokenization and Alignment.} Several approaches integrate alignment considerations into tokenization. \citet{chung-gildea-2009-unsupervised} propose using word alignment between parallel sentences for Chinese word segmentation. While their approach shares similarities with our work through its foundation in word alignment, key differences exist: (1) they derive tokenization from alignment, whereas we compute tokenization directly with alignment as a by-product, and (2) they use an explicit hyperparameter to control tokenized sequence compression, while in our method, compression emerges naturally from the algorithm.

\citet{deguchi-etal-2020-bilingual} developed a machine translation-specific tokenization method that selects subword segmentations of parallel sentences to maximize unigram language model probability while maintaining similar length. This approach aims at better efficiency and reaches better text compression without sacrificing tokenization quality, but does not optimize for semantic overlap.

\paragraph{Word Alignment Methods.} The word alignment field includes statistical approaches such as the IBM models \cite{brown-ibm} and Eflomal \cite{Ostling2016efmaral}, as well as neural network-based methods like Awesome Align \cite{dou2021word}. These tools focus on the alignment task rather than integrating it with tokenization.

\section{Alignable Tokenization}\label{sec:method}

For cross-lingually alignable tokenization, we assume a fixed tokenizer for the source language and access to parallel data between the source and target languages. 
The goal is to derive a target-language tokenization such that subwords in both languages are semantically aligned. 
Moreover, we require that it is possible to reconstruct the original text by simply concatenating the tokens (and removing some special characters).
We adopt a probabilistic formulation similar to the Unigram tokenizer, but condition token probabilities on the fixed source language tokenization:
\begin{equation}\label{eq:formulation}
\mathcal{L}(T, S) = \argmax_{\Tok} \sum_{t \in \Tok(T)} -\log p(t \mid S)
\end{equation}
where $\Tok$ is a function that splits the target-language sequence $T$ into tokens, and $S$ is a source-language sequence encoded as tokens. $T$ is the translation of $S$. The objective is to find target-language character spans that align with source-language tokens.
% \MP{Why $\mathcal{L}(t, S)$ and not $\mathcal{L}(T, S)$?}

Estimating $p(t \mid S)$ directly is intractable. We simplify it by treating the source sentence as a bag of tokens and computing the probability as:
\begin{equation}\label{eq:conditional}
p(t \mid S) = \frac{p(t, S)}{p(S)} \approx \frac{\sum\limits_{s_i \in S } c(t, s_i)}{\sum\limits_{t_j \in \vtgt} \sum\limits_{s_k \in S} c(t_j, s_k)}
\end{equation}
where $c(t, s)$ counts the co-occurrences of tokens $t$ and $s$ in sequence pairs in a corpus containing parallel sentences, and $\vtgt$ represents the target vocabulary.

% \jl{Here, we should say, how we do the actual segmentation because the next paragraph says that we segment the data.}
%

Given $p(t \mid S)$, we find a segmentation that maximizes the overall probability using the Unigram model’s dynamic programming algorithm. 
Initially, the vocabulary $\vtgt$ contains all character spans from the training data (up to a fixed length), and $c(t, s)$ is estimated based on all the possible tokens in the target language and tokens in the source language.
% \MP{This should be clarified. Are the spans overlapping? So why do we call $t$ a token and not a substring?}
% (up to a fixed length) in the target language and tokens in the source language.
At every iteration, we update it by computing the expected number of co-occurrences and using only the target tokens currently in the vocabulary $\vtgt$. 
For a particular training example $T$, this is proportional to the probability of observing the prefix ($T_{:i}$), the token itself ($T_{i:j}$), and the suffix ($T_{j:}$). 
Then, the amount is distributed across the source tokens, so that the contribution of a pair of tokens $(T_{i:j}, s)$ from the training example $(T, S)$ is the following:
% \gv{I think I should also explain how p(tgt[:i]) is computed}
% \MP{Yes. tgt has not been defined.}
\begin{equation}\label{eq:expected}
    c_{\text{sample}}(T_{i:j}, s) = \dfrac{p(T_{i:j} \mid S) ~~ p(T_{:i} \mid S) ~~ p(T_{j:} \mid S)}{\operatorname{length}(S)}
\end{equation}

Then, these quantities are accumulated to obtain the updated count table.

We experiment also with an alternative training method similar to expectation maximization, where we iterate the following two steps: First, after initializing the
% \MP{Why "the" when no table has been mentioned so far?}
table $c(t, s)$, we use it to tokenize the text; Second, we use the tokenized text to update the table by increasing the count of the tokens that appear in it. 
However, with this method, tokens that do not appear during the first iteration are never counted and so they are immediately removed from the vocabulary. 
For this reason, in our experiments, we will compare both methods, but focus mostly on the former one.
% will use the other training algorithm for our experiments.

With either training method, we initialize the target vocabulary with all character spans up to a fixed length. Similarly to the unigram model, we reduce the vocabulary iteratively, always after adjusting the unigram probabilities. We keep the subwords with the highest mutual information with the source tokens until the desired vocabulary size is reached. In this way, we can penalize pairs of tokens where one of them is rare while the other is frequent and that appear together by chance.
Single characters are always kept in the vocabulary.

% \gv{In practice I simplify this a bit more so that it uses c(t,s)}
\begin{equation} 
    I(t,\vsrc) = \sum_{s \in \vsrc} p(t,s) \log\dfrac{p(t,s)}{p(t)p(s)}
\end{equation}

To reduce the memory requirements and speed up the training, we pretokenize the input sentences and use Eflomal \cite{Ostling2016efmaral} to align the words. Then, each pair of aligned words is used as a training example instead of the full sentences.
Although we do not experiment with languages without white spaces, this step can be skipped entirely or adapted to such languages by using a different pre-tokenization method.

Token alignment probabilities between two tokens $t$ and $s$ can be computed as:
\begin{equation}\label{eq:align}
p(t \mid s) = \dfrac{c(t, s)}{\sum\limits_{t_i \in \vtgt}c(t_i, s)}
\end{equation}
For a given target sequence, we consider only tokens that are substrings of the target sequence in the denominator instead of the entire vocabulary $\vtgt$.

This formulation requires both source and target sequences for target tokenization. Alternatively, only the target sequence can be used by estimating $p(t)$ via marginalization:
\begin{equation}\label{eq:unpaired}
p(t) = \sum_{s \in \vsrc} p(t, s) = \dfrac{\sum\limits_{s_i \in \vsrc}c(t, s_i)}{\sum\limits_{t_j \in \vtrg}\sum\limits_{s_k \in \vsrc} c(t_j, s_k)}
\end{equation}
where $\vsrc$ is the vocabulary of the fixed source tokenizer.
This resembles an unconditional Unigram tokenizer but with tokens counted differently. 
Alternatively, following \citet{libovicky-helcl-2024-lexically}, we can use the tokenized text to distill a bigram model.

The simplified pseudo-code for training our tokenizer is shown in Appendix~\ref{app:code}.

\section{Experiments} \label{sec:experiments}
First, we evaluate our model intrinsically and then on two tasks: machine translation, since it requires parallel data, and language modeling to investigate its performance without parallel data.

We focus on the following language pairs:

\paragraph{French (fra) \& Italian (ita).} 

Both languages are high resources (Tier 5 and 4 according to \citealp{joshi-etal-2020-state}) from the same family and use the same alphabet.

\paragraph{Czech (ces) \& Ukrainian (ukr).}

Compared to the previous pair, this is a less-resourced language pair (Tier 4 and 3). The languages are from the same family but use different scripts.

\paragraph{Italian (ita) \& Maltese (mlt).}

They differ in families but share the same script. Maltese, a low-resource Semitic language (Tier 2), has complex morphology with infixes but shows Italian influence due to geographical proximity.

\paragraph{German (deu) \& Upper Sorbian (hsb).}

Both languages are spoken in Germany, but they come from different families. German is a high-resource language (Tier 5), while Upper Sorbian is a low-resource Slavic language (Tier 1).

For French-Italian and Czech-Ukrainian, we train the tokenizers with 100k, 500k, and 1M examples. 
The data is from NLLB \cite{nllbteam2022languageleftbehindscaling}, which contains 47M examples for French-Italian and 4M for Czech-Ukrainian. 
For Italian-Maltese, we use 100k examples from MultiParaCrawl \cite{banon-etal-2020-paracrawl}, which totals 483k examples.
For German-Upper Sorbian, we use 60k examples from WMT2020 \cite{libovicky-fraser-2021-findings}.

We use Flores \cite{nllbteam2022languageleftbehindscaling} for evaluating the tokenizers, with the exception of German-Upper Sorbian, which is evaluated on the WMT2020 test set.

\subsection{Intrinsic Evaluation}

We compare our tokenizers against Unigram models from SentencePiece trained on identical data with matching vocabulary sizes (8k, 16k, 32k). These baseline models also serve as the source tokenizers for training our conditional tokenizers.
We use the following notation: \spsrc and \sptrg refer to SentencePiece tokenizers for source and target languages, respectively (e.g., for the pair \lang{Czech}{Ukrainian}, \spsrc is trained on Czech, \sptrg on Ukrainian), while \psp refers to \textit{Paired SentencePiece}. We also evaluate two variants: \psp trained with Expectation Maximization (\pspem), and a version that tokenizes only target sequences without source context (\usp). Note that \psp and \usp share identical parameters (the co-occurrence table $c(t, s)$) but differ in their tokenization procedures.

We assess tokenization quality using the following metrics:

\paragraph{Parity ($\downarrow$).} This measures the ratio of tokens produced by our tokenizer in the target language to those produced by the reference tokenizer in the source language \cite{petrov_language_2023}. Optimal tokenization should yield similar sequence lengths across languages.

\paragraph{Fertility ($\downarrow$).} This measures the average number of tokens per word \cite{rust_how_2021}. Lower fertility (minimum 1.0) indicates that words remain coherent semantic units. %However, following Zipf's law, using only word-level tokens would result in many rare tokens with poorly trained embeddings, therefore, we include another metric to address this issue. \jl{What do you mean, another metric? Some Renyi efficiency that you removed?} \gv{Yes, I forgot to change the sentence}

\begin{figure*}[t]
    \centering
    \includegraphics[width=0.95\linewidth]{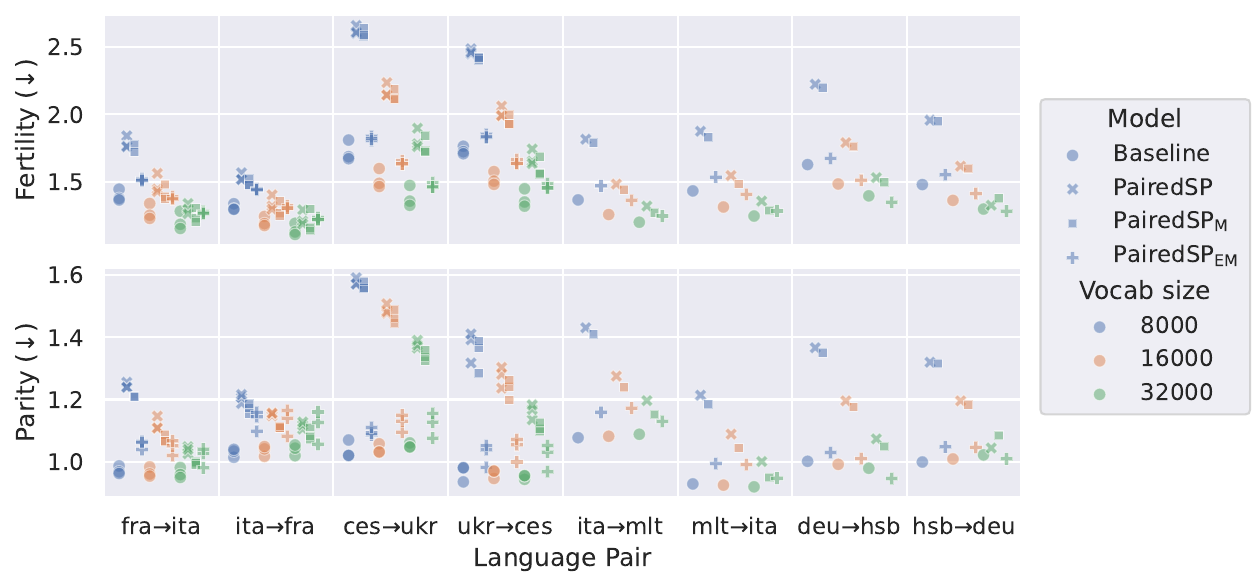}
    \caption{Fertility and parity scores of the tokenizers on the different language pairs, subdivided by vocabulary size (color) and model (shape). There is an outlier (\usp \lang{fra}{ita} 8k vocabulary size) that is not shown for clarity: it has 5.61 fertility and 3.95 parity.}
    \label{fig:tok}
\end{figure*}

\begin{table}[t]
    \caption{Parity scores of the different tokenizers when trained with the largest training set available for the language pairs.}
    \label{tab:parity}
    \centering
    \footnotesize
    \setlength{\tabcolsep}{3pt}
    \begin{tabular}{ll ccc ccc}
        \toprule
        \multicolumn{8}{c}{\textbf{Parity ($\downarrow$)}} \\        
        \midrule
        Size & Model & 8k & 16k & 32k & 8k & 16k & 32k \\
        \midrule
        & & \multicolumn{3}{c}{\textbf{fra $\rightarrow$ ita}} & 
            \multicolumn{3}{c}{\textbf{ita $\rightarrow$ fra}} \\
            \cmidrule(lr){3-5} \cmidrule(lr){6-8}
        \multirow{4}*{1m} & \psp & 1.24 & 1.11 & 1.04 & 1.22 & 1.16 & 1.13 \\
         & \usp & 3.95 & 1.07 & 0.99 & 1.19 & 1.11 & 1.08 \\
         & \pspem & 1.06 & 1.07 & 1.04 & 1.16 & 1.16 & 1.16 \\
        & \sptgt & \textbf{0.96} & \textbf{0.95} & \textbf{0.95} & \textbf{1.04} & \textbf{1.05} & \textbf{1.05} \\
        \midrule
        & & \multicolumn{3}{c}{\textbf{ces $\rightarrow$ ukr}} & 
            \multicolumn{3}{c}{\textbf{ukr $\rightarrow$ ces}} \\
            \cmidrule(lr){3-5} \cmidrule(lr){6-8}
        \multirow{4}*{1m} & \psp & 1.59 & 1.51 & 1.39 & 1.41 & 1.30 & 1.18 \\
         & \usp & 1.58 & 1.49 & 1.36 & 1.39 & 1.26 & 1.13 \\
         & \pspem & 1.11 & 1.15 & 1.16 & 1.05 & 1.07 & 1.05 \\
        & \sptgt & \textbf{1.02} & \textbf{1.03} & \textbf{1.05} & \textbf{0.98} & \textbf{0.97} & \textbf{0.95} \\
        \midrule
        & & \multicolumn{3}{c}{\textbf{ita $\rightarrow$ mlt}} & 
            \multicolumn{3}{c}{\textbf{mlt $\rightarrow$ ita}} \\
            \cmidrule(lr){3-5} \cmidrule(lr){6-8}
        \multirow{4}*{100k} & \psp & 1.43 & 1.28 & 1.20 & 1.21 & 1.09 & 1.00 \\
         & \usp & 1.41 & 1.24 & 1.15 & 1.19 & 1.04 & 0.95 \\
         & \pspem & 1.16 & 1.17 & 1.13 & 0.99 & 0.99 & 0.95 \\
        & \sptgt & \textbf{1.08} & \textbf{1.08} & \textbf{1.09} & \textbf{0.93} & \textbf{0.92} & \textbf{0.92} \\
        \midrule
        & & \multicolumn{3}{c}{\textbf{deu $\rightarrow$ hsb}} & 
            \multicolumn{3}{c}{\textbf{hsb $\rightarrow$ deu}} \\
            \cmidrule(lr){3-5} \cmidrule(lr){6-8}
        \multirow{4}*{60k} & \psp & 1.37 & 1.20 & 1.07 & 1.32 & 1.20 & 1.04 \\
         & \usp & 1.35 & 1.18 & 1.05 & 1.32 & 1.18 & 1.08 \\
         & \pspem & 1.03 & 1.01 & \textbf{0.95} & 1.05 & 1.05 & \textbf{1.01} \\
        & \sptgt & \textbf{1.00} & \textbf{0.99} & 0.98 & \textbf{1.00} & \textbf{1.01} & 1.02 \\
        \bottomrule
    \end{tabular}
\end{table}

\begin{table}[t]
    \caption{Fertility scores of the different tokenisers when trained with the largest training set available for the language pairs.}
    \label{tab:fertility}
    \centering
    \footnotesize
    \setlength{\tabcolsep}{3pt}
    \begin{tabular}{ll ccc ccc}
        \toprule
        \multicolumn{8}{c}{\textbf{Fertility ($\downarrow$)}} \\        
        \midrule
        Size & Model & 8k & 16k & 32k & 8k & 16k & 32k \\
        \midrule
        & & \multicolumn{3}{c}{\textbf{fra $\rightarrow$ ita}} & 
            \multicolumn{3}{c}{\textbf{ita $\rightarrow$ fra}} \\
            \cmidrule(lr){3-5} \cmidrule(lr){6-8}
        \multirow{4}*{1m} & \psp & 1.76 & 1.43 & 1.26 & 1.52 & 1.30 & 1.19 \\
         & \usp & 5.61 & 1.37 & 1.20 & 1.48 & 1.25 & 1.14 \\
         & \pspem & 1.51 & 1.38 & 1.27 & 1.45 & 1.31 & 1.22 \\
        & \sptgt & \textbf{1.37} & \textbf{1.23} & \textbf{1.15} & \textbf{1.30} & \textbf{1.18} & \textbf{1.11} \\
        \midrule
        & & \multicolumn{3}{c}{\textbf{ces $\rightarrow$ ukr}} & 
            \multicolumn{3}{c}{\textbf{ukr $\rightarrow$ ces}} \\
            \cmidrule(lr){3-5} \cmidrule(lr){6-8}
        \multirow{4}*{1m} & \psp & 2.61 & 2.14 & 1.76 & 2.46 & 1.99 & 1.64 \\
         & \usp & 2.59 & 2.12 & 1.72 & 2.42 & 1.93 & 1.56 \\
         & \pspem & 1.82 & 1.63 & 1.46 & 1.83 & 1.64 & 1.45 \\
        & \sptgt & \textbf{1.67} & \textbf{1.47} & \textbf{1.33} & \textbf{1.71} & \textbf{1.48} & \textbf{1.32} \\
        \midrule
        & & \multicolumn{3}{c}{\textbf{ita $\rightarrow$ mlt}} & 
            \multicolumn{3}{c}{\textbf{mlt $\rightarrow$ ita}} \\
            \cmidrule(lr){3-5} \cmidrule(lr){6-8}
        \multirow{4}*{100k} & \psp & 1.82 & 1.48 & 1.32 & 1.87 & 1.55 & 1.36 \\
         & \usp & 1.79 & 1.44 & 1.27 & 1.83 & 1.48 & 1.29 \\
         & \pspem & 1.47 & 1.36 & 1.25 & 1.53 & 1.41 & 1.28 \\
        & \sptgt & \textbf{1.37} & \textbf{1.26} & \textbf{1.20} & \textbf{1.43} & \textbf{1.31} & \textbf{1.25} \\
        \midrule
        & & \multicolumn{3}{c}{\textbf{deu $\rightarrow$ hsb}} & 
            \multicolumn{3}{c}{\textbf{hsb $\rightarrow$ deu}} \\
            \cmidrule(lr){3-5} \cmidrule(lr){6-8}
        \multirow{4}*{60k} & \psp & 2.22 & 1.79 & 1.53 & 1.96 & 1.62 & 1.33 \\
         & \usp & 2.20 & 1.76 & 1.50 & 1.95 & 1.60 & 1.38 \\
         & \pspem & 1.67 & 1.51 & \textbf{1.35} & 1.55 & 1.41 & \textbf{1.28} \\
        & \sptgt & \textbf{1.63} & \textbf{1.48} & 1.40 & \textbf{1.48} & \textbf{1.36} & 1.30 \\
        \bottomrule
    \end{tabular}
\end{table}

For alignment quality assessment, we first get the token alignment on the test data using Eflomal and  we compare \psp and \sptgt using:

\paragraph{One-to-one ($\uparrow$).} Following \citet{hammerl-etal-2025-beyond}, this measures the proportion of source tokens that have exactly one aligned target token which is also aligned to exactly one token. We measure this on the source side due to its fixed tokenization.
%
% \gv{Not really clear. There can be only that many 1-1 links, if the text is well aligned but over tokenized, we would have a lower score if we measure it on the target side}

% \paragraph{Eflomal-score ($\downarrow$).} Used by  \citet{hammerl-etal-2025-beyond}, it is the “maximum unnormalized log-probability of links in the last sampling iteration” \cite{vazquez-etal-2019-university}. \TODO \gv{This is taken from Kathy's paper}

\paragraph{Unaligned ($\downarrow$).} It is the portion of source tokens that are not aligned to any target tokens. As for \textit{One-to-one}, we measure it on the source sequence. 

We tokenize the dev tests with both \sptgt and \psp, then mark the tokens to recognize which tokenizer produced them. After joining the two sets, we train Eflomal to align this set in the target language to the one in the source language, tokenized by \spsrc. 
We prepare the test sets in the same way, and we use them to compute the alignment metrics with the Eflomal priors computed on the dev tests.
In this way, Eflomal can align sentences tokenized by either model, and we can compare the metrics computed for both tokenizers.

% \paragraph{R\'enyi efficiency.} It is an entropy-based measure that quantifies deviation from a uniform distribution. \citet{zouhar_tokenization_2023} show this metric correlates well with BLEU scores \cite{papineni_bleu_2002} in machine translation.

\begin{figure*}[t]
    \centering
    \includegraphics[width=0.95\linewidth]{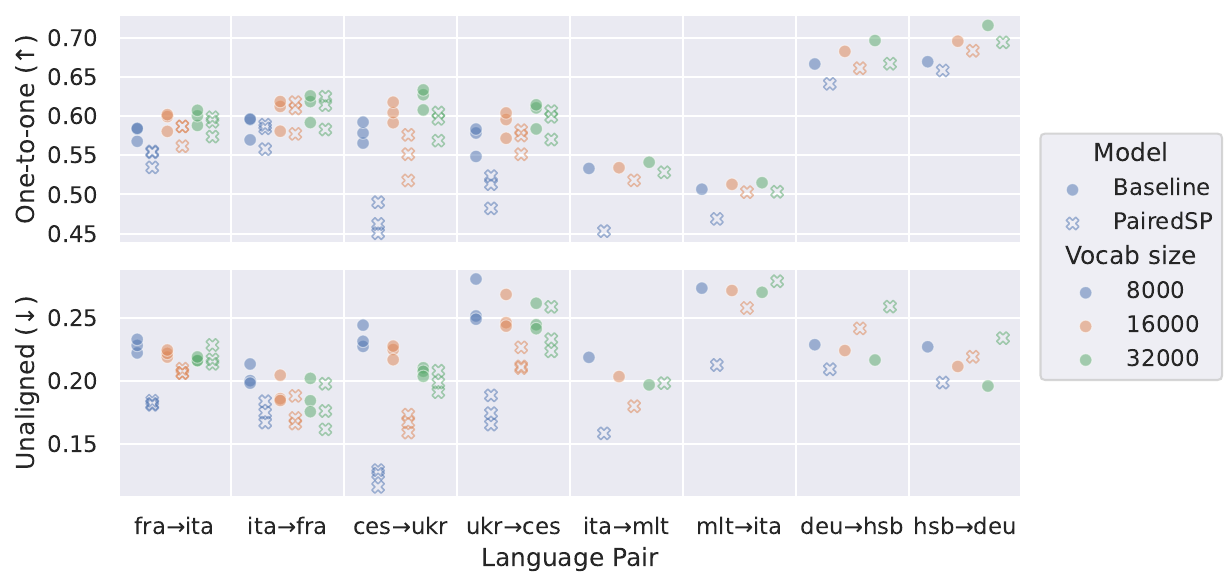}
    \caption{Alignment scores of the tokenizers on the different language pairs subdivided by vocabulary size.}
    \label{fig:align}
\end{figure*}

\begin{table}[t]
    \caption{One-to-one scores of the tokenizers trained on the largest training set available for each language.}
    \label{tab:one-to-one}
    \centering
    \footnotesize
    \setlength{\tabcolsep}{3pt}
    \begin{tabular}{ll ccc ccc}
        \toprule
        \multicolumn{8}{c}{\textbf{One-to-one ($\uparrow$)}} \\        
        \midrule
        Size & Model & 8k & 16k & 32k & 8k & 16k & 32k \\
        \midrule
        & & \multicolumn{3}{c}{\textbf{fra $\rightarrow$ ita}} & 
            \multicolumn{3}{c}{\textbf{ita $\rightarrow$ fra}} \\
            \cmidrule(lr){3-5} \cmidrule(lr){6-8}
        \multirow{2}*{1m} & \psp & 0.55 & 0.59 & 0.60 & 0.59 & \textbf{0.62} & 0.62 \\
        & \sptgt & \textbf{0.58} & \textbf{0.60} & \textbf{0.61} & \textbf{0.60} & \textbf{0.62} &\textbf{ 0.63} \\
        \midrule
        & & \multicolumn{3}{c}{\textbf{ces $\rightarrow$ ukr}} & 
            \multicolumn{3}{c}{\textbf{ukr $\rightarrow$ ces}} \\
            \cmidrule(lr){3-5} \cmidrule(lr){6-8}
        \multirow{2}*{1m} & \psp & 0.49 & 0.58 & 0.60 & 0.52 & 0.58 & 0.61 \\
        & \sptgt & \textbf{0.59} & \textbf{0.62} & \textbf{0.63} & \textbf{0.58} & \textbf{0.60} & \textbf{0.61} \\
        \midrule
        & & \multicolumn{3}{c}{\textbf{ita $\rightarrow$ mlt}} & 
            \multicolumn{3}{c}{\textbf{mlt $\rightarrow$ ita}} \\
            \cmidrule(lr){3-5} \cmidrule(lr){6-8}
        \multirow{2}*{100k} & \psp & 0.45 & 0.52 & 0.53 & 0.47 & 0.50 & 0.50 \\
        & \sptgt & \textbf{0.53} & \textbf{0.53} & \textbf{0.54} & \textbf{0.51} & \textbf{0.51} & \textbf{0.51} \\
        \midrule
        & & \multicolumn{3}{c}{\textbf{deu $\rightarrow$ hsb}} & 
            \multicolumn{3}{c}{\textbf{hsb $\rightarrow$ deu}} \\
            \cmidrule(lr){3-5} \cmidrule(lr){6-8}
        \multirow{2}*{60k} & \psp & 0.64 & 0.66 & 0.67 & 0.66 & 0.68 & 0.69 \\
        & \sptgt & \textbf{0.67} & \textbf{0.68} & \textbf{0.70} & \textbf{0.67} & \textbf{0.70} & \textbf{0.72} \\
        \bottomrule
    \end{tabular}
\end{table}

\begin{table}[t]
    \caption{Unaligned scores of the tokenizers trained on the largest training set available for each language.}
    \label{tab:unaligned}
    \centering
    \footnotesize
    \setlength{\tabcolsep}{3pt}
    \begin{tabular}{ll ccc ccc}
        \toprule
        \multicolumn{8}{c}{\textbf{Unaligned ($\downarrow$)}} \\        
        \midrule
        Size & Model & 8k & 16k & 32k & 8k & 16k & 32k \\
        \midrule
        & & \multicolumn{3}{c}{\textbf{fra $\rightarrow$ ita}} & 
            \multicolumn{3}{c}{\textbf{ita $\rightarrow$ fra}} \\
            \cmidrule(lr){3-5} \cmidrule(lr){6-8}
        \multirow{2}*{1m} & \psp & \textbf{0.18} & \textbf{0.21} & \textbf{0.21} & \textbf{0.17} & \textbf{0.17} & \textbf{0.16} \\
        & \sptgt & 0.23 & 0.22 & 0.22 & 0.20 & 0.18 & 0.18 \\
        \midrule
        & & \multicolumn{3}{c}{\textbf{ces $\rightarrow$ ukr}} & 
            \multicolumn{3}{c}{\textbf{ukr $\rightarrow$ ces}} \\
            \cmidrule(lr){3-5} \cmidrule(lr){6-8}
        \multirow{2}*{1m} & \psp & \textbf{0.13} & \textbf{0.17} & \textbf{0.19} & \textbf{0.17} & \textbf{0.21} & \textbf{0.22} \\
        & \sptgt & 0.23 & 0.22 & 0.20 & 0.25 & 0.24 & 0.24 \\
        \midrule
        & & \multicolumn{3}{c}{\textbf{ita $\rightarrow$ mlt}} & 
            \multicolumn{3}{c}{\textbf{mlt $\rightarrow$ ita}} \\
            \cmidrule(lr){3-5} \cmidrule(lr){6-8}
        \multirow{2}*{100k} & \psp & \textbf{0.16} & \textbf{0.18} & \textbf{0.20} & \textbf{0.21} & \textbf{0.26} & 0.28 \\
        & \sptgt & 0.22 & 0.20 & 0.20 & 0.27 & 0.27 & \textbf{0.27} \\
        \midrule
        & & \multicolumn{3}{c}{\textbf{deu $\rightarrow$ hsb}} & 
            \multicolumn{3}{c}{\textbf{hsb $\rightarrow$ deu}} \\
            \cmidrule(lr){3-5} \cmidrule(lr){6-8}
        \multirow{2}*{60k} & \psp & \textbf{0.21} & 0.24 & 0.26 & \textbf{0.20} & 0.22 & 0.23 \\
        & \sptgt & 0.23 & \textbf{0.22} & \textbf{0.22} & 0.23 & \textbf{0.21} & \textbf{0.20 }\\
        \bottomrule
    \end{tabular}
\end{table}

\subsection{Machine Translation}

We evaluate our tokenizer on machine translation, hypothesizing that improved token correspondence between languages should simplify MT model training by making the task more similar to token-level translation rather than complex sequence-to-sequence mapping.

We use the same language pairs and tokenizers as in intrinsic evaluation, testing three vocabulary sizes with the largest available training set for each language pair. Our experimental setup uses \spsrc for input tokenization and \psp for output tokenization, with \sptrg replacing \psp as the baseline.

\begin{figure*}[t]
    \centering
    \includegraphics[width=0.95\linewidth]{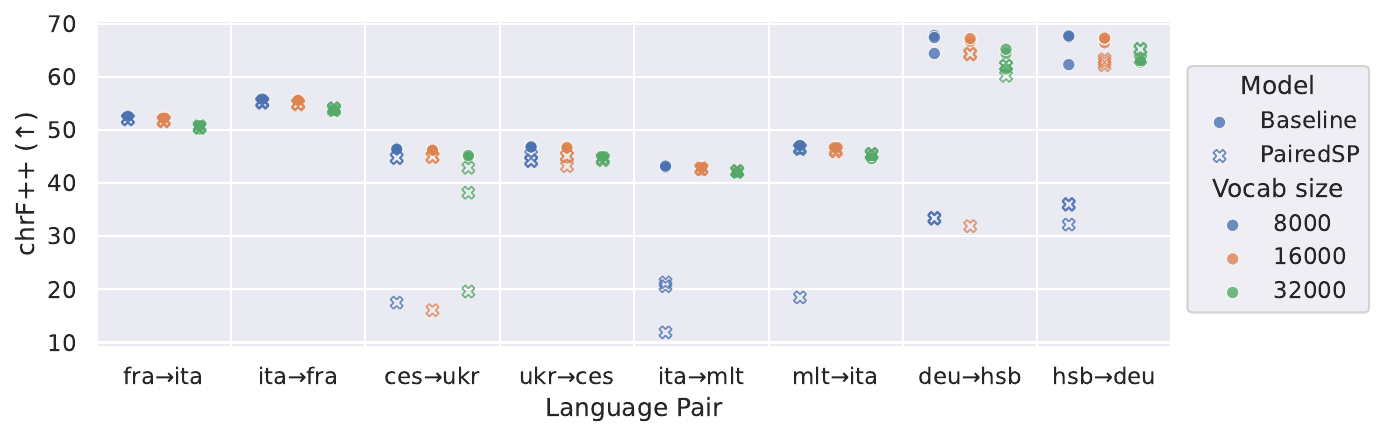}
    \caption{chrF++ ($\uparrow$) scores on the different language pairs and vocabulary sizes. For most pairs, the baseline has higher scores than \psp and lower variance.}
    \label{fig:chrf}
\end{figure*}

\begin{table}[t]
    \caption{Average chrF++ scores on the different language pairs and vocabulary sizes.}
    \label{tab:chrF++}
    \centering
    \footnotesize
    \setlength{\tabcolsep}{3pt}
    \begin{tabular}{l ccc ccc}
        \toprule
        \multicolumn{7}{c}{\textbf{chrF++ ($\uparrow$)}} \\
        \midrule
        Model & 8k & 16k & 32k & 8k & 16k & 32k \\
        \midrule
        & \multicolumn{3}{c}{\textbf{fra $\rightarrow$ ita}} & \multicolumn{3}{c}{\textbf{ita $\rightarrow$ fra}} \\ 
            \cmidrule(lr){2-4} \cmidrule(lr){5-7}
        \spsrc + \psp & 52.0 & 51.7 & 50.5 &
            55.2 & 54.9 & \textbf{53.9} \\
        \spsrc + \sptgt & \textbf{52.5} & \textbf{52.3} & \textbf{50.6} &
            \textbf{55.8} & \textbf{55.5} & 53.7 \\
        \midrule
        & \multicolumn{3}{c}{\textbf{ces $\rightarrow$ ukr}} &
            \multicolumn{3}{c}{\textbf{ukr $\rightarrow$ ces}} \\ \cmidrule(lr){2-4} \cmidrule(lr){5-7}
        \spsrc + \psp & 35.7 & 35.3 & 33.6 &
            44.6 & 44.4 & 44.5 \\
        \spsrc + \sptgt & \textbf{46.4} & \textbf{46.1} & \textbf{45.0} &
            \textbf{46.9} & \textbf{46.6} & \textbf{44.9} \\
        \midrule
        & \multicolumn{3}{c}{\textbf{ita $\rightarrow$ mlt}} &
            \multicolumn{3}{c}{\textbf{mlt $\rightarrow$ ita}} \\ \cmidrule(lr){2-4} \cmidrule(lr){5-7}
        \spsrc + \psp & 17.9 & 42.7 & \textbf{42.2} &
            37.2 & 46.1 & \textbf{45.4} \\
        \spsrc + \sptgt & \textbf{43.0} & \textbf{42.8} & 42.0 &
            \textbf{46.8} & \textbf{46.5} & 45.1 \\
        \midrule
        & \multicolumn{3}{c}{\textbf{deu $\rightarrow$ hsb}} &
            \multicolumn{3}{c}{\textbf{hsb $\rightarrow$ deu}} \\ \cmidrule(lr){2-4} \cmidrule(lr){5-7}
        \spsrc + \psp & 33.4 & 53.5 & 61.4 &
            34.7 & 62.8 & \textbf{64.6} \\
        \spsrc + \sptgt & \textbf{66.5} & \textbf{67.0} & \textbf{63.7} &
            \textbf{65.8} & \textbf{67.0} & 62.9 \\
        \bottomrule
    \end{tabular}    
\end{table}

We train the models using Marian \cite{mariannmt} with the Transformer-base architecture \cite{vaswani_attention_2017} (hyperparameter details in Appendix~\ref{app:hyper}). Each model undergoes 1M training updates using data from NLLB, MultiParaCrawl, or WMT2020, depending on the language pair.

We evaluate models using chrF++ on Flores test sets (and WMT2020 test set for German-Upper Sorbian), and additionally report BLEU, TER from SacreBLEU \cite{post-2018-call}, and COMET scores \cite{rei-etal-2020-comet}. Complete details are provided in Appendices~\ref{app:mt-metrics} and \ref{app:mt}.

\subsection{Language Modeling}

% \gv{Double check that the perplexity is computed correctly and that the values make sense}

Finally, we evaluate our tokenizer in a setting without access to parallel data during inference. We train small GPT-2-like models \cite{radford2019language} (91M to 110M parameters depending on the vocabulary size) from scratch using the HuggingFace implementation on the target language of each language pair (hyperparameter details in Appendix~\ref{app:lm-hyper}).
% \jl{Is there a name for the model size? Small? Tiny?} \gv{It should be like this one openai-community/gpt2, with 124M parameters}

We compare two settings, monolingual and bilingual, and each model is trained on a fixed number of examples (2M) to ensure fair comparison. We compare \usp against \sptgt as the baseline. Importantly, while monolingual models observe only monolingual data during training, the \usp tokenizer was trained with cross-lingual support from \spsrc, allowing us to assess whether cross-lingual tokenizer training benefits monolingual language modeling.
Models are tested only on the target language, and we use perplexity per byte to compare models with different vocabularies.

\begin{figure*}
    \centering
    \includegraphics[width=0.95\linewidth]{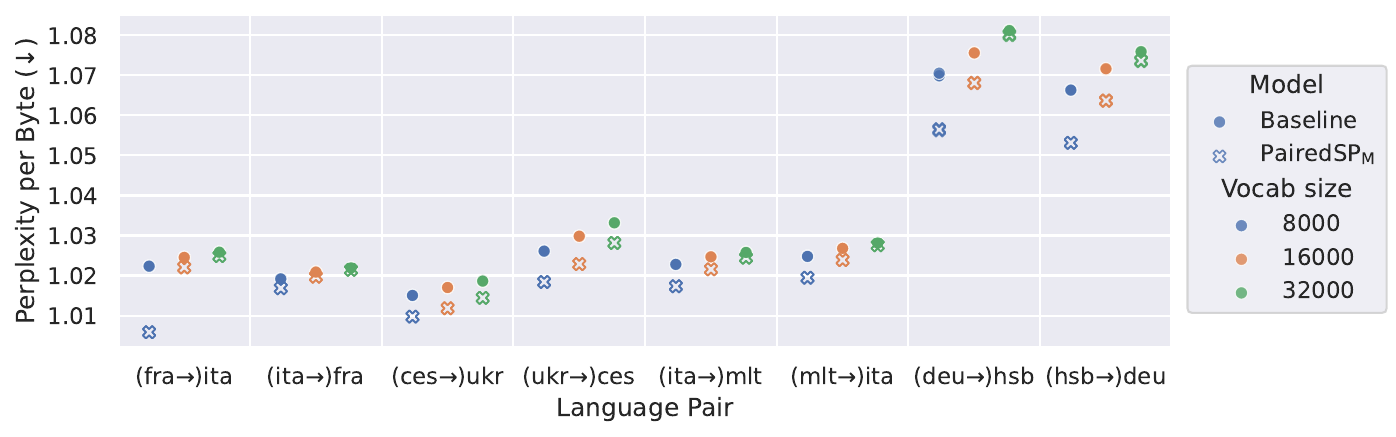}
    \caption{Perplexity per byte of bilingual language models trained on the different languages, subdivided by vocabulary size. \usp has generally better scores than the baseline, and the models have less variance than in the MT task. In parentheses, there is the source language used to train \psp.}
    \label{fig:ppl}
\end{figure*}

\begin{table}[t]
    \centering
    \footnotesize
    \setlength{\tabcolsep}{3pt}
        \caption{Average Perplexity per Byte ($\downarrow$) scores on the different language pairs and vocabulary size. }
    \label{tab:eval_ppl_bytes}
    \begin{tabular}{ll ccc ccc}
        \toprule
        \multicolumn{8}{c}{\textbf{Perplexity per Byte ($\downarrow$)}} \\
        \midrule
        Model & Setting & 8k & 16k & 32k & 8k & 16k & 32k \\
        \midrule
        & & \multicolumn{3}{c}{\textbf{(fra $\rightarrow$) ita}} & \multicolumn{3}{c}{\textbf{(ita $\rightarrow$) fra}} \\ 
            \cmidrule(lr){3-5} \cmidrule(lr){6-8}
        \usp & Mono & \textbf{1.006} & \textbf{1.021} & \textbf{1.024} &
            \textbf{1.016} & \textbf{1.019} & \textbf{1.020} \\
        \usp & Bi & 1.006 & 1.022 & 1.025 &
            1.017 & 1.020 & 1.021 \\
        \sptgt & Mono & 1.021 & 1.023 & 1.025 &
            1.018 & 1.020 & 1.021 \\
        \sptgt & Bi & 1.022 & 1.025 & 1.026 &
            1.019 & 1.021 & 1.022 \\
        \midrule
        & & \multicolumn{3}{c}{\textbf{(ces $\rightarrow$) ukr}} &
            \multicolumn{3}{c}{\textbf{(ukr $\rightarrow$) ces}} \\ 
            \cmidrule(lr){3-5} \cmidrule(lr){6-8}
        \usp & Mono & \textbf{1.009} & \textbf{1.011} & \textbf{1.014} &
            \textbf{1.017} & \textbf{1.022} & \textbf{1.027} \\
        \usp & Bi & 1.010 & 1.012 & 1.014 &
            1.018 & 1.023 & 1.028 \\
        \sptgt & Mono & 1.014 & 1.016 & 1.018 &
            1.025 & 1.028 & 1.031 \\
        \sptgt & Bi & 1.015 & 1.017 & 1.019 &
            1.026 & 1.030 & 1.033 \\
        \midrule
        & & \multicolumn{3}{c}{\textbf{(ita $\rightarrow$) mlt}} &
            \multicolumn{3}{c}{\textbf{(mlt $\rightarrow$) ita}} \\ 
            \cmidrule(lr){3-5} \cmidrule(lr){6-8}
        \usp & Mono & 1.018 & 1.023 & 1.026 &
            1.020 & 1.025 & 1.029 \\
        \usp & Bi & \textbf{1.017} & \textbf{1.022} & \textbf{1.024} &
            \textbf{1.019} & \textbf{1.024} & \textbf{1.028} \\
        \sptgt & Mono & 1.024 & 1.026 & 1.027 &
            1.026 & 1.028 & 1.030 \\
        \sptgt & Bi & 1.023 & 1.025 & 1.026 &
            1.025 & 1.027 & 1.028 \\
        \midrule
        & & \multicolumn{3}{c}{\textbf{(deu $\rightarrow$) hsb}} &
            \multicolumn{3}{c}{\textbf{(hsb $\rightarrow$) deu}} \\ 
            \cmidrule(lr){3-5} \cmidrule(lr){6-8}
        \usp & Mono & 1.063 & 1.074 & 1.085 &
            1.059 & 1.069 & 1.078 \\
        \usp & Bi & \textbf{1.056} & \textbf{1.068} & \textbf{1.080} &
            \textbf{1.053} & \textbf{1.064} & \textbf{1.074} \\
        \sptgt & Mono & 1.077 & 1.081 & 1.085 &
            1.073 & 1.078 & 1.080 \\        
        \sptgt & Bi & 1.070 & 1.076 & 1.081 &
            1.066 & 1.072 & 1.076 \\
        \bottomrule
    \end{tabular}
\end{table}

\section{Results \& Discussion}\label{sec:results}
\subsection{Intrinsic Evaluation}

Figure~\ref{fig:tok} presents the intrinsic tokenization metrics.
The baseline consistently outperforms \psp and its variants on both parity and fertility metrics, though this difference diminishes with larger vocabulary sizes.
\usp shows comparable performance to \psp, indicating that marginalization does not substantially impact performance in most cases. However, there is one notable failure case: with French-Italian using 1M training examples and 8k vocabulary size, \usp produces only single-character tokens, resulting in fertility and parity scores of 5.61 and 3.95, respectively.

As expected, larger vocabulary sizes generally improve these metrics. Contrary to our expectations, \pspem performs better than \psp despite not updating counts for rare tokens, though it still underperforms the baseline. Additionally, as shown in Tables~\ref{tab:parity} and~\ref{tab:fertility}, \usp outperforms \psp. We hypothesize that this occurs because \usp's probability estimation more closely resembles that of \sptgt.

We observe similar patterns in the one-to-one alignment metric. \psp shows improvement over the baseline on the unaligned metric, indicating that it leaves fewer source tokens without target alignments. While larger vocabulary sizes improve the one-to-one metric consistently, they improve the unaligned metric only for the baseline, possibly due to increased difficulty in estimating the $c(t,s)$ table that is quadratically bigger compared to the unconditional probability $p(t)$ in the Unigram model.

\subsection{Machine Translation}

Figure~\ref{fig:chrf} shows chrF++ scores for machine translation models across language pairs and vocabulary sizes. The baseline consistently outperforms our model. For some language pairs like French-Italian, the difference is minimal (0.33 chrF++ on average), while for others like Czech-Ukrainian, it is more substantial (6.31 chrF++ on average). Vocabulary size appears to have minimal effect on results, with a slight decrease in scores for larger vocabularies given the same number of training steps.

Table~\ref{tab:chrF++} shows that \psp outperforms the baseline in only four cases on average. Notably, Czech-Ukrainian shows the largest performance decrease when using \psp, though this pattern does not hold in the reverse direction.
%
%These results align with the observations of \citet{hammerl-etal-2025-beyond}, who found that one-to-one alignment positively correlates with machine translation performance.

% In Table~\ref{tab:chrF++} we can see how, on average, \psp is better than the baseline only in four cases. Also, the pair \lang{Czech}{Ukrainian} has the largest decrease in performance when using \psp, but this does not happen in the other direction.
%
% These results reflect the observations of \citet{hammerl-etal-2025-beyond}, where one-to-one positively correlates with machine translation results. \gv{(double check that mt is in the paper) xnli, dep. tagging, ner}\jl{nope, MT is not there.}
%
Furthermore, scores with our tokenizer exhibit much higher variance than the baseline (except for French-Italian), suggesting that this tokenization approach may be less reliable than standard SentencePiece.

\subsection{Language Modeling}

Figure~\ref{fig:ppl} demonstrates that \usp achieves improved perplexity per byte across all language pairs and vocabulary sizes compared to the baseline. Interestingly, this improvement does not correlate with tokenization scores: the \usp model with the worst intrinsic evaluation scores achieves the lowest perplexity in language modeling.
Table~\ref{tab:eval_ppl_bytes} shows that bilingual training with both the source and target language improves the perplexity per byte on the target language in low-resource languages.

\section{Conclusions} \label{sec:conclusion}
% comment on scaling, but bad for low resource
%  Our results present a mixed picture: while the intrinsic evaluation shows that our conditional tokenizer maintains statistical properties comparable to standard unigram tokenizers, we do not observe consistet improvements in machine translation quality. However, we do find notable perplexity reductions in language modeling tasks, suggesting potential benefits for specific applications
% 28M fertility, 4M 1-1, parity is sometimes negative
%In this paper, we presented a novel tokenization method that aims to improve tokenization with the use of parallel data. 
%However, our results do not improve in the intrinsic evaluation or in the machine learning task, we observe perplexity reduction in the language modeling task.
%We hypothesize the difference in the performance between \psp and the Unigram model is cuase but the increase complexity of the problem: $p(t)$ scales linearly with the vocabulary size, while $p(t|S)$ quadratically, but the training data is still the same, thus this tokenizer is not suited for low-resource languages.
%We expect to need around 28M example to match Unigram fertily and 4M for one-to-one. \gv{Assuming linear improvement. For parity we often get a negative size.}
%Further research is needed to research a data-efficient tokenization method with good sub-word alignability.

We presented a novel tokenization method that leverages parallel data to improve cross-lingual token alignment. Our approach extends the unigram tokenization framework by conditioning target token probabilities on source language tokens, with the goal of achieving better semantic alignment between languages.

Our experimental evaluation reveals mixed results across different tasks and metrics. While our method does not consistently improve intrinsic tokenization metrics or machine translation quality compared to standard unigram tokenizers, we observe consistent perplexity reductions in language modeling tasks. 

We hypothesize that the performance gap between our approach and standard unigram tokenization stems from the increased memory complexity of the underlying estimation problem: while a table storing $p(t)$ scales linearly with vocabulary size, $p(t \mid S)$ scales quadratically, yet the available training data remains the same. This scaling issue may particularly impact low-resource languages, contrary to our initial motivation.

Based on our observations, we estimate that approximately 28M examples would be required to match unigram fertility and 4M examples for comparable one-to-one alignment performance. These requirements may limit the practical applicability of our approach, especially for the low-resource scenarios where improved tokenization is most needed.

Future work should explore more data-efficient methods for learning cross-lingually aligned tokenizations. Potential directions include investigating alternative parameterizations that scale more favorably with vocabulary size, exploring pre-training strategies that leverage multilingual representations, or developing hybrid approaches that combine the benefits of both conditional and unconditional tokenization methods.

\section*{Impact Statement}
% \gv{Required by ICML}
The aim of this paper is to contribute to diminishing the gap between high- and low-resource languages.
However, it investigates only a limited number of languages and from a limited geographical area.
We do not see any direct ethical risk related to this work.

\section*{Acknowledgements}

We thank Martin Popel for comments on the draft of this paper.
This research was supported by the Czech Science Foundation project 25-16242S.
 
\bibliography{anthology,custom}
\bibliographystyle{icml2025}

\newpage
\appendix
\onecolumn

\section{Examples}
The Tables~\ref{tab:tok-examples}~\ref{tab:mt-examples} and~\ref{tab:lm-examples} show respectively tokenization, machine translation, and language modelling examples.

\begin{table}[ht]
    \captionof{table}{Tokenization examples from the different models on two specific settings. Tokens are separated by a white space. "▁" denotes a white space in the original sentence, which can be reconstructed by concatenating the tokens.}
    \label{tab:tok-examples}
    % \smallskip
    \begin{tabularx}{0.95\linewidth}{l|X}
        \toprule
        \multicolumn{2}{c}{\textbf{\lang{Fra}{Ita} 8k, vocabulary, 1M training set}}\\            
         \spsrc (ref.) & ▁« ▁Nous ▁avons ▁à ▁présent ▁des ▁souri s ▁de ▁4 ▁mois [...] \\
         \sptgt & ▁" ▁Abbiam o ▁top i ▁di ▁quattro ▁mesi [...] \\
         \psp & ▁" ▁Abbiamo ▁to p i ▁di ▁quattro ▁mesi [...]\\
         \pspem & ▁" ▁Abbiamo ▁ to pi ▁di ▁quattro ▁mesi [...]\\
         \usp & ▁ " ▁ A b b i a m o ▁ t o p i ▁ d i ▁ q u a t t r o ▁ m e s I [...]\\            
        \midrule
        \multicolumn{2}{c}{\textbf{\lang{Ces}{Ukr}, 32k vocabulary, 500k training set}}\\
         \spsrc (ref.) & \foreignlanguage{ukrainian}{▁" ▁Зараз ▁у ▁нас ▁є ▁4 ▁- ▁місячні ▁миші} \\
         \sptgt & ▁„ ▁Nyn í ▁má me ▁čtyř měsíční ▁myši\\
         \psp & ▁„ ▁Nyní ▁máme ▁čtyř mě s í č ní ▁myši \\
         \pspem & ▁„ ▁Nyní ▁máme ▁čtyř měsíční ▁myš i\\
         \usp & ▁„ ▁Nyní ▁máme ▁čtyř mě s í č ní ▁myši\\ 
         \midrule
         \multicolumn{2}{c}{\textit{"We (now) have four-month-old mice [...]}}\\
         \bottomrule
    \end{tabularx}
\end{table}

\begin{table*}[ht]
    \captionof{table}{Machine translation examples from two specific settings. The output of the model is shown tokenized.}\label{tab:mt-examples}
    \begin{tabularx}{0.95\linewidth}{l|>{\raggedright\arraybackslash}X}
            \toprule
            \multicolumn{2}{c}{\textbf{\lang{Czech}{Ukrainian}, 8k vocabulary}}\\           
             Source & „Nyní máme čtyřměsíční myši bez cukrovky, které ji dříve měly,“ dodal.\\
             Target & \foreignlanguage{ukrainian}{"Зараз у нас є 4-місячні миші, в яких немає діабету і які мали діабет раніше," — додав він.} \\
             \spsrc+\sptgt & \foreignlanguage{ukrainian}{▁" ▁Зараз ▁у ▁нас ▁чотири місячн а ▁ми ша ▁без ▁діабет у ▁, ▁яка ▁раніше ▁була ▁у ▁неї ▁" ▁, ▁- ▁додав ▁він ▁.}\\
             \spsrc+\psp & \foreignlanguage{ukrainian}{▁" ▁Зараз ▁у ▁нас ▁є ▁чотири м і с я ч н і ▁ми ш і ▁без ▁д і а б е т у ▁, ▁які ▁раніше ▁мали ▁" ▁, ▁- ▁додав ▁він ▁.}\\
            \midrule
            \multicolumn{2}{c}{\textit{"We now have 4-month-old mice that are non-diabetic that used to be diabetic," he added.}}\\
            \midrule
            \multicolumn{2}{c}{\textbf{\lang{Upper Sorbian}{German}, 32k vocabulary}}\\
             Source & To njepłaći jenož za naše měšćanske zarjadnišća.\\
             Target & Das gilt nicht nur für unsere städtischen Verwaltungen.\\
             \spsrc+\sptgt & ▁Dies ▁g ilt ▁nicht ▁ nur ▁für ▁unsere ▁unsere ▁städtische n ▁Einrichtung en ▁.\\
             \spsrc+\psp & ▁Das ▁gilt ▁nicht ▁nur ▁für ▁unsere ▁Stadtverwaltung ▁.\\        
             \midrule
             \multicolumn{2}{c}{\textit{This does not only apply to our municipal administrations.}}\\
             \bottomrule
    \end{tabularx}
\end{table*}

\begin{table*}[ht]
    \captionof{table}{Language modelling examples from two specific settings. The output of the model is shown tokenized. The prediction of model, computed in an autoregressive way, is shown in bold.}\label{tab:lm-examples}    
    \begin{tabularx}{0.95\linewidth}{l|>{\raggedright\arraybackslash}X}
            \toprule
            \multicolumn{2}{c}{\textbf{\lang{(Italian}{)French}, 16k vocabulary}}\\           
             \sptgt & ▁« ▁Nous ▁avons ▁à ▁présent ▁des ▁souris ▁de ▁4 ▁mois \textbf{▁qui ▁ont ▁été ▁infecté es ▁par ▁le ▁virus ▁de ▁la}\\
             \psp & ▁« ▁Nous ▁avons ▁à ▁présent ▁des ▁souris ▁de ▁4 ▁mois \textbf{▁et ▁des ▁souris ▁de ▁plus ▁de ▁6 ▁mois ▁, ▁mais} \\
            \midrule
            \multicolumn{2}{c}{\textbf{\lang{(German}{)Upper Sorbian}, 32k vocabulary}}\\
             \sptgt & ▁To ▁njepłaći ▁je nož ▁za ▁naše ▁měšćanske ▁zarjadnišća ▁. </s> \textbf{e ▁, ▁ kotrež ▁ma my ▁tu ▁ja ra ▁dołh}\\
             \psp & ▁To ▁njepłaći ▁jenož ▁za ▁naše ▁měšćanske ▁zarjadnišća ▁. </s> </s> \textbf{▁, ▁kotrež ▁wustawki ▁Domowiny ▁, ▁kotryž ▁je ▁za ł o}\\
             \bottomrule
    \end{tabularx}
    % \smallskip    
\end{table*}

\section{Tokenizer preprocessing}
We use a character coverage of 1.0 and normalize the text with NFKC to reduce differences with the SentencePiece implementation. 
Moreover, we prepend whitespace to punctuation characters and to the beginning of the sentence. Then, whitespaces are replaced with U+2581.

The relevant SentencePiece settings are the following:
\begin{itemize}[noitemsep]
    \item character coverage: 1.0 
    \item shrinking\_factor: 0.75
    \item num\_sub\_iterations: 2 
    \item allow\_whitespace\_only\_pieces: true
    \item byte\_fallback: true
\end{itemize}
We use equivalent settings for \psp.

\section{Metrics for the Intrinsic Evaluation}\label{app:metrics}
This is a list of additional metrics we considered for the intrinsic evaluation for the tokenization task:
\begin{description}
    \item[Single Character tokens.]  We count the proportion of tokens in a sequence that are just a single character.
    \item[Vocabulary usage.] We compute the proportion of tokens in the vocabulary that is actually used on the test set.
    \item[Vocabulary overlap.] This is the portion of vocabulary that overlaps between our tokenizer and the reference one. Both tokenizers are trained on the same languages.
    \item[Length ratio with respect to \sptrg on target text.] Given parallel sequences, we take the ratio between the length in tokens of the target text with our tokenizer and the reference length.
    \item[R\'enyi efficiency ratio with respect to \sptrg.] This is analogous to the length ratio but for the R\'enyi efficiency, which is an entropy-based measure that quantifies deviation from a uniform distribution. \citet{zouhar_tokenization_2023} show that this metric correlates well with BLEU scores \cite{papineni_bleu_2002} in machine translation.
    \item[Begin of word.] We count the proportion of tokens in the vocabulary that represent the beginning of a word.
\end{description}

For the alignment task, we compute the additional metrics:
\begin{description}
    \item[Eflomal score.] \citet{hammerl-etal-2025-beyond} show that this correlates with cross-lingual transfer. It measures the "maximum unnormalized log-probability of links in the last sampling iteration" \cite{vazquez-etal-2019-university}.
\end{description}

\section{SacreBleu and COMET}\label{app:mt-metrics}
We use the following settings for SacreBleu:\\
\texttt{BLEU|nrefs:1|case:mixed|eff:no|tok:13a|smooth:exp|version:2.4.2}\\
\texttt{chrF2++|nrefs:1|case:mixed|eff:yes|nc:6|nw:2|space:no|version:2.4.2}\\
\texttt{TER|nrefs:1|case:lc|tok:tercom|norm:no|punct:yes|asian:no|version:2.4.2}

And \texttt{Unbabel/wmt22-comet-da} for computing COMET. Note that this model is not trained for Maltese or Upper Sorbian.

\section{Tokenizer Pseudo-code}\label{app:code}
Algorithm~\ref{alg:train} summarizes the training loop. Many details regarding the settings of the tokenizer (e.g., number of iterations, character coverage, ...) are left out for simplicity.

\begin{algorithm}[H]
    \caption{Training algorithm.}
    \label{alg:train}
    \begin{algorithmic}
        \FUNCTION{\textsc{Train}($src, trg$)}
            \INPUT{$src$: list of tokenized source sentences, $trg$: list of target sentences}
            \OUTPUT{$c$: co-occurence table, $\mathcal{V}$: vocabulary}
            \STATE $c \gets 0$
            \STATE $\mathcal{V} \gets \{\}$
            \STATE \COMMENT{Initialize the co-occurrence table}
            \FORALL{$(S, T) \in (src, trg)$}
                \FORALL{$(t, s) \in \CALL{Spans}{T} \times S$}
                    \STATE $c(t, s)\gets c(t,s)+1$
                    \STATE $\mathcal{V} \gets \mathcal{V} \cup \{t\}$
                \ENDFOR
            \ENDFOR
            \STATE \COMMENT{Training iterations}
            \FOR{$i \gets 1 $ \textbf{to} $n_{iterations}$}
                \STATE $c \gets \CALL{Count}{c, \mathcal{V}, src, trg}$
                \IF{$i \mod n_{subiterations}$ == 0}
                    \STATE \COMMENT{Remove the target tokens with the lowest $I(t, \vsrc)$}
                    \STATE $\mathcal{V} \gets \CALL{Prune}{V}$
                \ENDIF
            \ENDFOR
            \STATE \COMMENT{Update the table with the final vocabulary.}
            \STATE $c \gets \CALL{Count}{c, \mathcal{V}, src, trg}$
            \RETURN $c$, $\mathcal{V}$
        \ENDFUNCTION
        \vspace{0.5em}
        \FUNCTION{\textsc{Count}($c, \mathcal{V}, src, trg$)}            
            \STATE $c_{new} \gets 0$
            \FORALL{$(T) \in (src, trg)$}
                \FORALL{$(t, s) \in \CALL{Spans}{T} \times T$}
                    \IF{$t \in \mathcal{V}$}
                        \STATE $pref \gets \CALL{Prefix}{T, t}$
                        \STATE $suff \gets \CALL{Suffix}{T, t}$
                        \STATE \COMMENT{\CALL{Score}{$\dots$} computes the conditional probability of a token given a sentence and the co-occurrence table.}
                        \STATE \COMMENT{\CALL{Score$_{tok}$}{$\dots$} is similar, marginalize over the possible tokenizations of the prefix or suffix.}
                        \STATE $c_{new}(t, s)\gets c_{new}(t,s)+ \dfrac{\CALL{Score}{c, t, S} \CALL{Score$_{tok}$}{c, pref, S} \CALL{Score$_{tok}$}{c, suff, S}}{\operatorname{length}(src)}$
                    \ENDIF
                \ENDFOR
            \ENDFOR
            \RETURN $c_{new}$
        \ENDFUNCTION
        % update table
                % for src trg
                    % for span in trg
                    % c(t,s) += prefix * scores * suffix / len src
            
    \end{algorithmic}
\end{algorithm}
\begin{algorithm*}
\vspace{.5em}
\caption{Tokenization algorithm, adapted from Unigram.}\label{alg:tok}
\begin{algorithmic}
    \FUNCTION{\textsc{Tokenize}($c, src, trg$)}
        \STATE \COMMENT{$\sim$ Forward pass from Unigram}
        \FORALL{$(S, T) \in (src, trg)$}         
            \STATE $best \gets [0, -\infty, ...]$
            \STATE $sizes \gets [0, 0, ...]$
            \STATE \COMMENT{Iterate over the spans starting from $i$}
            \FOR{$i\gets 1, \operatorname{length}(T)+1$}
                \FOR{$j \gets i-1, -1$}
                    \STATE $t \gets T[j:i]$
                    \STATE \COMMENT{$t$ is in the vocabulary}
                    \IF{$c(t, :) \neq 0$} 
                        \STATE $score \gets p(t \mid S)$ 
                        \STATE \COMMENT{Store the loss and size of the token}
                        \IF{$(best[j] + score) > best[i]$} 
                            \STATE $best[i] \gets best[j] + score$ 
                            \STATE $sizes[i] \gets i - j$ 
                        \ENDIF
                    \ENDIF
                \ENDFOR
            \ENDFOR
            \STATE \COMMENT{$\sim$ Backward pass from Unigram}
            \STATE $\mathcal{L} \gets best[-1]$
            \STATE $i \gets \ell(sizes)$
            \STATE $tok \gets []$
            \STATE \COMMENT{Add tokens with size from $sizes$}
            \WHILE{$i > 1$} 
                \STATE $next \gets i - sizes[i-1]$
                \STATE \CALL{Append}{$tok, T[next - 1: i - 1]$}
                \STATE $i \gets next$
            \ENDWHILE
            \STATE \textbf{yield} \CALL{Reverse}{$tok$}
        \ENDFOR
    \ENDFUNCTION
\end{algorithmic}
\end{algorithm*}

\section{Machine Translation Hyperparameters}\label{app:hyper}

\begin{table}[H]
    % \caption{Marian configuration.}
    \label{tab:mt-model}
    \centering
    \footnotesize
    \begin{tabular}{ll|ll}
        \toprule
        \multicolumn{2}{c}{\textbf{Model options}} & \multicolumn{2}{c}{\textbf{Validation options}}\\
        type & transformer & beam-size & 8 \\
        dim-emb & 512 & normalize & 0.6 \\
        enc-depth & 6 & valid-freq & 100ku \\
        dec-depth & 6 & \multirow{5}*{valid-metrics} & ce-mean-words \\
        tied-embeddings & true & & bleu \\
        transformer-heads & 8 & & perplexity \\
        transformer-dim-ffn & 2048 & & translation \\
        transformer-ffn-activation & relu & & chrf \\
        transformer-preprocess & "" & valid-mini-batch & 16 \\
        transformer-postprocess & dan\\
        transformer-dropout & 0.1 \\
        \midrule
        \multicolumn{4}{c}{\textbf{Training options}} \\
        cost-type & ce-mean-words & lr-warmup & 16000 \\
        max-length & 512 & lr-report & true \\
        mini-batch & 1000 & label-smoothing & 0.1 \\
        mini-batch-fit & true & clip-norm & 1 \\
        maxi-batch & 1000 & exponential-smoothing & 0.0001 \\
        optimizer-params & [0.9, 0.98, 1e-09] & disp-freq & 10ku \\
        sync-sgd & true & early-stopping & 10 \\
        learn-rate & 0.0003 & after & 1Mu \\
        lr-decay-inv-sqrt & 16000 & shuffle-in-ram & true \\
        \bottomrule
    \end{tabular}
\end{table}

\section{Language Modelling Hyperparameters}\label{app:lm-hyper}

\begin{table}[H]
    \label{tab:lm-model}
    \centering
    \footnotesize
    \begin{tabular}{ll|ll}
        \toprule
        \multicolumn{2}{c}{\textbf{Model options}} & \multicolumn{2}{c}{\textbf{Training options}} \\
        activation\_function & gelu\_new & per\_device\_train\_batch\_size & 64\\
        attn\_pdrop & 0.1 & per\_device\_eval\_batch\_size & 64\\
        embd\_pdrop & 0.1 & gradient\_accumulation\_steps & 8\\
        initializer\_range & 0.02 & max\_steps & 2\_000\_000 / 64 \\
        layer\_norm\_epsilon & 1e-05 & weight\_decay & 0.1\\
        model\_type & gpt2 & warmup\_steps & 1\_000\\
        n\_embd & 768 & lr\_scheduler\_type & cosine\\
        n\_head & 12 & learning\_rate & 5e-5\\
        n\_inner & null & fp16 & True\\
        n\_layer & 12\\
        n\_positions & 512\\
        reorder\_and\_upcast\_attn & false\\
        resid\_pdrop & 0.1\\
        scale\_attn\_by\_inverse\_layer\_idx & false\\
        scale\_attn\_weights & true\\
        transformers\_version & 4.51.2\\
        use\_cache & true\\
        vocab\_size & 8000/16000/32000\\
        \bottomrule
    \end{tabular}
\end{table}

\twocolumn[\section{Intrinsic Evaluation}\label{app:intrinsic}]
\begin{table}[H]
    \caption{Additional parity scores from the models trained on the smaller training sets.}
    \label{tab:parity-app}
    \centering
    \footnotesize
    \setlength{\tabcolsep}{3pt}
    \begin{tabular}{ll ccc ccc}
        \toprule
        \multicolumn{8}{c}{\textbf{Parity ($\downarrow$)}} \\        
        \midrule
        Size & Model & 8k & 16k & 32k & 8k & 16k & 32k \\
        \midrule
        & & \multicolumn{3}{c}{\textbf{Fra $\rightarrow$ Ita}} & 
            \multicolumn{3}{c}{\textbf{Ita $\rightarrow$ Fra}} \\
            \cmidrule(lr){3-5} \cmidrule(lr){6-8}
        \multirow{4}*{100k} & \psp & 1.26 & 1.15 & 1.03 & 1.19 & 1.15 & 1.10 \\
         & \usp & 1.21 & 1.09 & 1.00 & 1.15 & 1.11 & 1.11 \\
         & \pspem & 1.04 & 1.02 & 0.98 & 1.10 & 1.08 & 1.06 \\
        & \sptgt & 0.99 & 0.98 & 0.98 & 1.01 & 1.02 & 1.02 \\
        \midrule
        \multirow{4}*{500k} & \psp & 1.24 & 1.11 & 1.05 & 1.21 & 1.15 & 1.12 \\
         & \usp & 1.21 & 1.07 & 0.99 & 1.17 & 1.11 & 1.07 \\
         & \pspem & 1.07 & 1.05 & 1.03 & 1.14 & 1.14 & 1.13 \\
        & \sptgt & 0.97 & 0.96 & 0.96 & 1.03 & 1.04 & 1.04 \\
        \midrule    
        & & \multicolumn{3}{c}{\textbf{Ces $\rightarrow$ Ukr}} & 
            \multicolumn{3}{c}{\textbf{Ukr $\rightarrow$ Ces}} \\
            \cmidrule(lr){3-5} \cmidrule(lr){6-8}
        \multirow{4}*{100k} & \psp & 1.57 & 1.48 & 1.37 & 1.32 & 1.24 & 1.13 \\
         & \usp & 1.56 & 1.45 & 1.33 & 1.28 & 1.20 & 1.10 \\
         & \pspem & 1.09 & 1.10 & 1.08 & 0.98 & 1.00 & 0.97 \\
        & \sptgt & 1.07 & 1.06 & 1.06 & 0.93 & 0.95 & 0.94 \\
        \midrule
        \multirow{4}*{500k} & \psp & 1.57 & 1.48 & 1.38 & 1.39 & 1.28 & 1.17 \\
         & \usp & 1.56 & 1.46 & 1.34 & 1.37 & 1.24 & 1.11 \\
         & \pspem & 1.09 & 1.13 & 1.13 & 1.04 & 1.05 & 1.03 \\
        & \sptgt & 1.02 & 1.03 & 1.05 & 0.98 & 0.97 & 0.95 \\        
        \bottomrule
    \end{tabular}
\end{table}

\begin{table}[H]
    \caption{Additional fertility scores from the models trained on the smaller training sets.}
    \label{tab:fertility-app}
    \centering
    \footnotesize
    \setlength{\tabcolsep}{3pt}
    \begin{tabular}{ll ccc ccc}
        \toprule
        \multicolumn{8}{c}{\textbf{Fertiliy ($\downarrow$)}} \\        
        \midrule
        Size & Model & 8k & 16k & 32k & 8k & 16k & 32k \\
        \midrule
        & & \multicolumn{3}{c}{\textbf{Fra $\rightarrow$ Ita}} & 
            \multicolumn{3}{c}{\textbf{Ita $\rightarrow$ Fra}} \\
            \cmidrule(lr){3-5} \cmidrule(lr){6-8}
        \multirow{4}*{100k} & \psp & 1.84 & 1.56 & 1.34 & 1.57 & 1.40 & 1.29 \\
         & \usp & 1.78 & 1.48 & 1.31 & 1.52 & 1.36 & 1.30 \\
         & \pspem & 1.52 & 1.39 & 1.28 & 1.45 & 1.32 & 1.24 \\
        & \sptgt & 1.45 & 1.34 & 1.28 & 1.34 & 1.24 & 1.19 \\
        \midrule
        \multirow{4}*{500k} & \psp & 1.77 & 1.45 & 1.30 & 1.52 & 1.32 & 1.21 \\
         & \usp & 1.72 & 1.39 & 1.23 & 1.47 & 1.27 & 1.16 \\
         & \pspem & 1.52 & 1.37 & 1.27 & 1.44 & 1.30 & 1.22 \\
        & \sptgt & 1.38 & 1.25 & 1.19 & 1.30 & 1.19 & 1.13 \\
        \midrule
        & & \multicolumn{3}{c}{\textbf{Ces $\rightarrow$ Ukr}} & 
            \multicolumn{3}{c}{\textbf{Ukr $\rightarrow$ Ces}} \\
            \cmidrule(lr){3-5} \cmidrule(lr){6-8}
        \multirow{4}*{100k} & \psp & 2.66 & 2.24 & 1.90 & 2.49 & 2.06 & 1.74 \\
         & \usp & 2.64 & 2.19 & 1.84 & 2.42 & 2.00 & 1.69 \\
         & \pspem & 1.84 & 1.66 & 1.49 & 1.85 & 1.67 & 1.49 \\
        & \sptgt & 1.81 & 1.60 & 1.47 & 1.76 & 1.58 & 1.45 \\
        \midrule
        \multirow{4}*{500k} & \psp & 2.60 & 2.14 & 1.78 & 2.45 & 1.99 & 1.65 \\
         & \usp & 2.58 & 2.11 & 1.73 & 2.40 & 1.93 & 1.57 \\
         & \pspem & 1.81 & 1.63 & 1.46 & 1.83 & 1.64 & 1.46 \\
        & \sptgt & 1.69 & 1.49 & 1.36 & 1.72 & 1.51 & 1.35 \\
        \bottomrule
    \end{tabular}
\end{table}

\begin{table}[H]
    \caption{Single Character tokens in the tokenized text.}
    \label{tab:single-char-micro}
    \centering
    \footnotesize
    \setlength{\tabcolsep}{3pt}
    \begin{tabular}{ll ccc ccc}
        \toprule
        \multicolumn{8}{c}{\textbf{Single Character  ($\downarrow$)}} \\        
        \midrule
        Size & Model & 8k & 16k & 32k & 8k & 16k & 32k \\
        \midrule
        & & \multicolumn{3}{c}{\textbf{Fra $\rightarrow$ Ita}} & 
            \multicolumn{3}{c}{\textbf{Ita $\rightarrow$ Fra}} \\
            \cmidrule(lr){3-5} \cmidrule(lr){6-8}
        \multirow{4}*{100k} & \psp & 0.34 & 0.23 & 0.13 & 0.23 & 0.17 & 0.13 \\
         & \usp & 0.31 & 0.18 & 0.10 & 0.21 & 0.14 & 0.13 \\
         & \pspem & 0.09 & 0.07 & 0.06 & 0.09 & 0.07 & 0.06 \\
        & \sptgt & 0.14 & 0.13 & 0.12 & 0.11 & 0.09 & 0.07 \\
        \midrule
        \multirow{4}*{500k} & \psp & 0.32 & 0.17 & 0.12 & 0.24 & 0.14 & 0.10 \\
         & \usp & 0.30 & 0.14 & 0.07 & 0.21 & 0.11 & 0.06 \\
         & \pspem & 0.09 & 0.07 & 0.06 & 0.10 & 0.07 & 0.06 \\
        & \sptgt & 0.09 & 0.08 & 0.07 & 0.09 & 0.06 & 0.05 \\
        \midrule
        \multirow{4}*{1M} & \psp & 0.32 & 0.17 & 0.11 & 0.25 & 0.14 & 0.09 \\
         & \usp & 1.00 & 0.13 & 0.07 & 0.23 & 0.10 & 0.05 \\
         & \pspem & 0.09 & 0.07 & 0.06 & 0.10 & 0.08 & 0.06 \\
        & \sptgt & 0.07 & 0.06 & 0.05 & 0.09 & 0.06 & 0.04 \\
        \midrule
        & & \multicolumn{3}{c}{\textbf{Ces $\rightarrow$ Ukr}} & 
            \multicolumn{3}{c}{\textbf{Ukr $\rightarrow$ Ces}} \\
            \cmidrule(lr){3-5} \cmidrule(lr){6-8}
        \multirow{4}*{100k} & \psp & 0.59 & 0.49 & 0.39 & 0.52 & 0.39 & 0.30 \\
         & \usp & 0.59 & 0.48 & 0.37 & 0.50 & 0.36 & 0.28 \\
         & \pspem & 0.12 & 0.10 & 0.09 & 0.13 & 0.11 & 0.09 \\
        & \sptgt & 0.24 & 0.19 & 0.16 & 0.22 & 0.19 & 0.16 \\
        \midrule
        \multirow{4}*{500k} & \psp & 0.59 & 0.48 & 0.36 & 0.53 & 0.39 & 0.28 \\
         & \usp & 0.58 & 0.47 & 0.34 & 0.51 & 0.36 & 0.23 \\
         & \pspem & 0.11 & 0.10 & 0.08 & 0.12 & 0.10 & 0.08 \\
        & \sptgt & 0.16 & 0.12 & 0.10 & 0.19 & 0.15 & 0.12 \\
        \midrule
        \multirow{4}*{1M} & \psp & 0.59 & 0.48 & 0.36 & 0.53 & 0.40 & 0.27 \\
         & \usp & 0.59 & 0.47 & 0.34 & 0.52 & 0.37 & 0.23 \\
         & \pspem & 0.11 & 0.10 & 0.08 & 0.13 & 0.10 & 0.08 \\
        & \sptgt & 0.15 & 0.11 & 0.09 & 0.18 & 0.14 & 0.11 \\
        \midrule
        & & \multicolumn{3}{c}{\textbf{Ita $\rightarrow$ Mlt}} & 
            \multicolumn{3}{c}{\textbf{Mlt $\rightarrow$ Ita}} \\
            \cmidrule(lr){3-5} \cmidrule(lr){6-8}
        \multirow{4}*{100k} & \psp & 0.41 & 0.24 & 0.16 & 0.38 & 0.24 & 0.16 \\
         & \usp & 0.41 & 0.22 & 0.13 & 0.36 & 0.21 & 0.12 \\
         & \pspem & 0.09 & 0.08 & 0.06 & 0.09 & 0.08 & 0.06 \\
        & \sptgt & 0.10 & 0.08 & 0.07 & 0.12 & 0.12 & 0.11 \\
        \midrule
        & & \multicolumn{3}{c}{\textbf{Deu $\rightarrow$ Hsb}} & 
            \multicolumn{3}{c}{\textbf{Hsb $\rightarrow$ Deu}} \\
            \cmidrule(lr){3-5} \cmidrule(lr){6-8}
        \multirow{4}*{60k} & \psp & 0.52 & 0.37 & 0.27 & 0.42 & 0.30 & 0.14 \\
         & \usp & 0.51 & 0.35 & 0.25 & 0.42 & 0.30 & 0.19 \\
         & \pspem & 0.10 & 0.08 & 0.07 & 0.09 & 0.08 & 0.05 \\
        & \sptgt & 0.20 & 0.18 & 0.15 & 0.19 & 0.16 & 0.12 \\
        \bottomrule
    \end{tabular}
\end{table}

\begin{table}[H]
    \caption{Vocabulary usage of the different tokenizers.}
    \label{tab:vocab-usage}
    \centering
    \footnotesize
    \setlength{\tabcolsep}{3pt}
    \begin{tabular}{ll ccc ccc}
        \toprule
        \multicolumn{8}{c}{\textbf{Vocabulary Usage  ($\uparrow$)}} \\        
        \midrule
        Size & Model & 8k & 16k & 32k & 8k & 16k & 32k \\
        \midrule
        & & \multicolumn{3}{c}{\textbf{Fra $\rightarrow$ Ita}} & 
            \multicolumn{3}{c}{\textbf{Ita $\rightarrow$ Fra}} \\
            \cmidrule(lr){3-5} \cmidrule(lr){6-8}
        \multirow{4}*{100k} & \psp & 0.43 & 0.28 & 0.18 & 0.43 & 0.28 & 0.17 \\
         & \usp & 0.43 & 0.29 & 0.17 & 0.43 & 0.28 & 0.16 \\
         & \pspem & 0.52 & 0.35 & 0.22 & 0.48 & 0.33 & 0.20 \\
        & \sptgt & 0.57 & 0.36 & 0.21 & 0.55 & 0.34 & 0.19 \\
        \midrule
        \multirow{4}*{500k} & \psp & 0.44 & 0.30 & 0.18 & 0.46 & 0.30 & 0.18 \\
         & \usp & 0.45 & 0.31 & 0.19 & 0.47 & 0.31 & 0.19 \\
         & \pspem & 0.52 & 0.36 & 0.21 & 0.50 & 0.34 & 0.20 \\
        & \sptgt & 0.57 & 0.38 & 0.21 & 0.56 & 0.36 & 0.20 \\
        \midrule
        \multirow{4}*{1M} & \psp & 0.45 & 0.31 & 0.19 & 0.46 & 0.31 & 0.19 \\
         & \usp & 0.01 & 0.31 & 0.19 & 0.47 & 0.32 & 0.19 \\
         & \pspem & 0.52 & 0.36 & 0.21 & 0.49 & 0.34 & 0.20 \\
        & \sptgt & 0.56 & 0.38 & 0.22 & 0.56 & 0.37 & 0.20 \\
        \midrule
        & & \multicolumn{3}{c}{\textbf{Ces $\rightarrow$ Ukr}} & 
            \multicolumn{3}{c}{\textbf{Ukr $\rightarrow$ Ces}} \\
            \cmidrule(lr){3-5} \cmidrule(lr){6-8}
        \multirow{4}*{100k} & \psp & 0.40 & 0.28 & 0.18 & 0.41 & 0.28 & 0.18 \\
         & \usp & 0.41 & 0.28 & 0.18 & 0.42 & 0.29 & 0.19 \\
         & \pspem & 0.54 & 0.38 & 0.24 & 0.54 & 0.38 & 0.24 \\
        & \sptgt & 0.56 & 0.38 & 0.23 & 0.58 & 0.39 & 0.23 \\
        \midrule
        \multirow{4}*{500k} & \psp & 0.42 & 0.30 & 0.19 & 0.42 & 0.30 & 0.20 \\
         & \usp & 0.43 & 0.30 & 0.19 & 0.43 & 0.31 & 0.20 \\
         & \pspem & 0.55 & 0.39 & 0.25 & 0.55 & 0.39 & 0.25 \\
        & \sptgt & 0.58 & 0.41 & 0.24 & 0.58 & 0.41 & 0.25 \\
        \midrule
        \multirow{4}*{1M} & \psp & 0.42 & 0.30 & 0.19 & 0.42 & 0.30 & 0.20 \\
         & \usp & 0.43 & 0.30 & 0.20 & 0.42 & 0.31 & 0.20 \\
         & \pspem & 0.53 & 0.39 & 0.25 & 0.54 & 0.39 & 0.25 \\
        & \sptgt & 0.58 & 0.42 & 0.25 & 0.58 & 0.41 & 0.25 \\
        \midrule
        & & \multicolumn{3}{c}{\textbf{Ita $\rightarrow$ Mlt}} & 
            \multicolumn{3}{c}{\textbf{Mlt $\rightarrow$ Ita}} \\
            \cmidrule(lr){3-5} \cmidrule(lr){6-8}
        \multirow{4}*{100k} & \psp & 0.39 & 0.27 & 0.16 & 0.42 & 0.28 & 0.18 \\
         & \usp & 0.39 & 0.27 & 0.17 & 0.42 & 0.29 & 0.18 \\
         & \pspem & 0.50 & 0.34 & 0.21 & 0.51 & 0.34 & 0.21 \\
        & \sptgt & 0.54 & 0.35 & 0.20 & 0.55 & 0.37 & 0.21 \\
        \midrule
        & & \multicolumn{3}{c}{\textbf{Deu $\rightarrow$ Hsb}} & 
            \multicolumn{3}{c}{\textbf{Hsb $\rightarrow$ Deu}} \\
            \cmidrule(lr){3-5} \cmidrule(lr){6-8}
        \multirow{4}*{60k} & \psp & 0.47 & 0.32 & 0.20 & 0.42 & 0.29 & 0.18 \\
         & \usp & 0.47 & 0.32 & 0.20 & 0.42 & 0.29 & 0.17 \\
         & \pspem & 0.55 & 0.38 & 0.24 & 0.50 & 0.33 & 0.21 \\
        & \sptgt & 0.60 & 0.39 & 0.23 & 0.58 & 0.37 & 0.21 \\
        \bottomrule
    \end{tabular}
\end{table}

\begin{table}[H]
    \caption{Vocabulary overlap with \sptgt.}
    \label{tab:vocab-overlap}
    \centering
    \footnotesize
    \setlength{\tabcolsep}{3pt}
    \begin{tabular}{ll ccc ccc}
        \toprule
        \multicolumn{8}{c}{\textbf{Vocabulary Overlap}} \\        
        \midrule
        Size & Model & 8k & 16k & 32k & 8k & 16k & 32k \\
        \midrule
        & & \multicolumn{3}{c}{\textbf{Fra $\rightarrow$ Ita}} & 
            \multicolumn{3}{c}{\textbf{Ita $\rightarrow$ Fra}} \\
            \cmidrule(lr){3-5} \cmidrule(lr){6-8}
        \multirow{4}*{100k} & \psp & 0.42 & 0.39 & 0.40 & 0.52 & 0.49 & 0.47 \\
         & \usp & 0.42 & 0.39 & 0.40 & 0.52 & 0.49 & 0.47 \\
         & \pspem & 0.34 & 0.29 & 0.31 & 0.34 & 0.30 & 0.32 \\
        \midrule
        \multirow{4}*{500k} & \psp & 0.58 & 0.56 & 0.50 & 0.64 & 0.62 & 0.57 \\
         & \usp & 0.58 & 0.56 & 0.50 & 0.64 & 0.62 & 0.57 \\
         & \pspem & 0.39 & 0.34 & 0.29 & 0.40 & 0.35 & 0.29 \\
        \midrule
        \multirow{4}*{1M} & \psp & 0.63 & 0.62 & 0.57 & 0.67 & 0.67 & 0.63 \\
         & \usp & 0.63 & 0.62 & 0.57 & 0.67 & 0.67 & 0.63 \\
         & \pspem & 0.42 & 0.35 & 0.30 & 0.42 & 0.36 & 0.30 \\
        \midrule
        & & \multicolumn{3}{c}{\textbf{Ces $\rightarrow$ Ukr}} & 
            \multicolumn{3}{c}{\textbf{Ukr $\rightarrow$ Ces}} \\
            \cmidrule(lr){3-5} \cmidrule(lr){6-8}
        \multirow{4}*{100k} & \psp & 0.39 & 0.38 & 0.34 & 0.46 & 0.43 & 0.39 \\
         & \usp & 0.39 & 0.38 & 0.34 & 0.46 & 0.43 & 0.39 \\
         & \pspem & 0.32 & 0.26 & 0.25 & 0.34 & 0.27 & 0.26 \\
        \midrule
        \multirow{4}*{500k} & \psp & 0.47 & 0.47 & 0.46 & 0.54 & 0.54 & 0.52 \\
         & \usp & 0.47 & 0.47 & 0.46 & 0.54 & 0.54 & 0.52 \\
         & \pspem & 0.41 & 0.33 & 0.29 & 0.43 & 0.34 & 0.31 \\
        \midrule
        \multirow{4}*{1M} & \psp & 0.51 & 0.51 & 0.51 & 0.56 & 0.57 & 0.57 \\
         & \usp & 0.51 & 0.51 & 0.51 & 0.56 & 0.57 & 0.57 \\
         & \pspem & 0.43 & 0.34 & 0.31 & 0.46 & 0.36 & 0.32 \\
        \midrule
        & & \multicolumn{3}{c}{\textbf{Ita $\rightarrow$ Mlt}} & 
            \multicolumn{3}{c}{\textbf{Mlt $\rightarrow$ Ita}} \\
            \cmidrule(lr){3-5} \cmidrule(lr){6-8}
        \multirow{4}*{100k} & \psp & 0.50 & 0.47 & 0.40 & 0.47 & 0.43 & 0.37 \\
         & \usp & 0.50 & 0.47 & 0.40 & 0.47 & 0.43 & 0.37 \\
         & \pspem & 0.38 & 0.30 & 0.29 & 0.35 & 0.28 & 0.29 \\
        \midrule
        & & \multicolumn{3}{c}{\textbf{Deu $\rightarrow$ Hsb}} & 
            \multicolumn{3}{c}{\textbf{Hsb $\rightarrow$ Deu}} \\
            \cmidrule(lr){3-5} \cmidrule(lr){6-8}
        \multirow{4}*{60k} & \psp & 0.33 & 0.33 & 0.31 & 0.42 & 0.39 & 0.35 \\
         & \usp & 0.33 & 0.33 & 0.31 & 0.42 & 0.39 & 0.35 \\
         & \pspem & 0.26 & 0.21 & 0.24 & 0.28 & 0.23 & 0.30 \\
        \bottomrule
    \end{tabular}
\end{table}

\begin{table}[H]
    \caption{Lnegth ratio between \psp (and derived models) and \sptrg.}
    \label{tab:length-ratio}
    \centering
    \footnotesize
    \setlength{\tabcolsep}{3pt}
    \begin{tabular}{ll ccc ccc}
        \toprule
        \multicolumn{8}{c}{\textbf{Length Ratio  ($\downarrow$)}} \\        
        \midrule
        Size & Model & 8k & 16k & 32k & 8k & 16k & 32k \\
        \midrule
        & & \multicolumn{3}{c}{\textbf{Fra $\rightarrow$ Ita}} & 
            \multicolumn{3}{c}{\textbf{Ita $\rightarrow$ Fra}} \\
            \cmidrule(lr){3-5} \cmidrule(lr){6-8}
        \multirow{4}*{100k} & \psp & 1.27 & 1.17 & 1.05 & 1.17 & 1.13 & 1.08 \\
         & \usp & 1.23 & 1.11 & 1.02 & 1.14 & 1.09 & 1.09 \\
         & \pspem & 1.05 & 1.04 & 1.00 & 1.08 & 1.06 & 1.04 \\
        \midrule
        \multirow{4}*{500k} & \psp & 1.28 & 1.15 & 1.09 & 1.17 & 1.11 & 1.07 \\
         & \usp & 1.25 & 1.11 & 1.04 & 1.13 & 1.07 & 1.03 \\
         & \pspem & 1.10 & 1.09 & 1.07 & 1.11 & 1.09 & 1.08 \\
        \midrule
        \multirow{4}*{1M} & \psp & 1.29 & 1.16 & 1.09 & 1.17 & 1.10 & 1.07 \\
         & \usp & 4.11 & 1.12 & 1.04 & 1.14 & 1.06 & 1.03 \\
         & \pspem & 1.10 & 1.12 & 1.10 & 1.11 & 1.11 & 1.10 \\
        \midrule
        & & \multicolumn{3}{c}{\textbf{Ces $\rightarrow$ Ukr}} & 
            \multicolumn{3}{c}{\textbf{Ukr $\rightarrow$ Ces}} \\
            \cmidrule(lr){3-5} \cmidrule(lr){6-8}
        \multirow{4}*{100k} & \psp & 1.47 & 1.40 & 1.29 & 1.41 & 1.31 & 1.20 \\
         & \usp & 1.46 & 1.37 & 1.25 & 1.37 & 1.27 & 1.16 \\
         & \pspem & 1.02 & 1.04 & 1.01 & 1.05 & 1.06 & 1.03 \\
        \midrule
        \multirow{4}*{500k} & \psp & 1.54 & 1.44 & 1.31 & 1.42 & 1.32 & 1.22 \\
         & \usp & 1.53 & 1.42 & 1.28 & 1.39 & 1.28 & 1.16 \\
         & \pspem & 1.07 & 1.10 & 1.08 & 1.06 & 1.09 & 1.08 \\
        \midrule
        \multirow{4}*{1M} & \psp & 1.56 & 1.46 & 1.33 & 1.44 & 1.34 & 1.24 \\
         & \usp & 1.55 & 1.44 & 1.30 & 1.42 & 1.30 & 1.18 \\
         & \pspem & 1.09 & 1.11 & 1.10 & 1.07 & 1.10 & 1.10 \\
        \midrule
        & & \multicolumn{3}{c}{\textbf{Ita $\rightarrow$ Mlt}} & 
            \multicolumn{3}{c}{\textbf{Mlt $\rightarrow$ Ita}} \\
            \cmidrule(lr){3-5} \cmidrule(lr){6-8}
        \multirow{4}*{100k} & \psp & 1.33 & 1.18 & 1.10 & 1.31 & 1.18 & 1.09 \\
         & \usp & 1.31 & 1.15 & 1.06 & 1.28 & 1.13 & 1.03 \\
         & \pspem & 1.08 & 1.08 & 1.04 & 1.07 & 1.07 & 1.03 \\
        \midrule
        & & \multicolumn{3}{c}{\textbf{Deu $\rightarrow$ Hsb}} & 
            \multicolumn{3}{c}{\textbf{Hsb $\rightarrow$ Deu}} \\
            \cmidrule(lr){3-5} \cmidrule(lr){6-8}
        \multirow{4}*{60k} & \psp & 1.37 & 1.21 & 1.10 & 1.32 & 1.19 & 1.02 \\
         & \usp & 1.35 & 1.19 & 1.07 & 1.32 & 1.17 & 1.06 \\
         & \pspem & 1.03 & 1.02 & 0.97 & 1.05 & 1.04 & 0.99 \\
        \bottomrule
    \end{tabular}
\end{table}

\begin{table}[H]
    \caption{Ratio between the R\'enyi efficiency of \psp (and derived models) and \sptgt}
    \label{tab:renyi-ratio}
    \centering
    \footnotesize
    \setlength{\tabcolsep}{3pt}
    \begin{tabular}{ll ccc ccc}
        \toprule
        \multicolumn{8}{c}{\textbf{Rényi Ratio  ($\uparrow$)}} \\        
        \midrule
        Size & Model & 8k & 16k & 32k & 8k & 16k & 32k \\
        \midrule
        & & \multicolumn{3}{c}{\textbf{Fra $\rightarrow$ Ita}} & 
            \multicolumn{3}{c}{\textbf{Ita $\rightarrow$ Fra}} \\
            \cmidrule(lr){3-5} \cmidrule(lr){6-8}
        \multirow{4}*{100k} & \psp & 0.92 & 0.95 & 0.99 & 0.95 & 0.97 & 0.98 \\
         & \usp & 0.93 & 0.97 & 1.00 & 0.96 & 0.98 & 0.98 \\
         & \pspem & 0.98 & 0.99 & 1.00 & 0.98 & 0.98 & 0.99 \\
        \midrule
        \multirow{4}*{500k} & \psp & 0.92 & 0.96 & 0.98 & 0.95 & 0.97 & 0.98 \\
         & \usp & 0.93 & 0.97 & 0.99 & 0.96 & 0.98 & 0.99 \\
         & \pspem & 0.97 & 0.98 & 0.98 & 0.97 & 0.98 & 0.98 \\
        \midrule
        \multirow{4}*{1M} & \psp & 0.92 & 0.96 & 0.98 & 0.95 & 0.98 & 0.99 \\
         & \usp & 0.47 & 0.97 & 0.99 & 0.96 & 0.99 & 0.99 \\
         & \pspem & 0.97 & 0.97 & 0.98 & 0.97 & 0.97 & 0.98 \\
        \midrule
        & & \multicolumn{3}{c}{\textbf{Ces $\rightarrow$ Ukr}} & 
            \multicolumn{3}{c}{\textbf{Ukr $\rightarrow$ Ces}} \\
            \cmidrule(lr){3-5} \cmidrule(lr){6-8}
        \multirow{4}*{100k} & \psp & 0.83 & 0.86 & 0.91 & 0.85 & 0.89 & 0.93 \\
         & \usp & 0.83 & 0.87 & 0.92 & 0.86 & 0.91 & 0.95 \\
         & \pspem & 0.99 & 0.99 & 1.00 & 0.98 & 0.98 & 0.99 \\
        \midrule
        \multirow{4}*{500k} & \psp & 0.82 & 0.86 & 0.91 & 0.85 & 0.89 & 0.93 \\
         & \usp & 0.82 & 0.87 & 0.92 & 0.86 & 0.90 & 0.95 \\
         & \pspem & 0.97 & 0.97 & 0.98 & 0.98 & 0.97 & 0.98 \\
        \midrule
        \multirow{4}*{1M} & \psp & 0.81 & 0.86 & 0.91 & 0.84 & 0.89 & 0.93 \\
         & \usp & 0.81 & 0.86 & 0.91 & 0.85 & 0.90 & 0.95 \\
         & \pspem & 0.97 & 0.96 & 0.97 & 0.97 & 0.96 & 0.97 \\
        \midrule
        & & \multicolumn{3}{c}{\textbf{Ita $\rightarrow$ Mlt}} & 
            \multicolumn{3}{c}{\textbf{Mlt $\rightarrow$ Ita}} \\
            \cmidrule(lr){3-5} \cmidrule(lr){6-8}
        \multirow{4}*{100k} & \psp & 0.90 & 0.95 & 0.97 & 0.91 & 0.95 & 0.98 \\
         & \usp & 0.90 & 0.96 & 0.98 & 0.92 & 0.97 & 0.99 \\
         & \pspem & 0.98 & 0.98 & 0.99 & 0.98 & 0.98 & 0.99 \\
        \midrule
        & & \multicolumn{3}{c}{\textbf{Deu $\rightarrow$ Hsb}} & 
            \multicolumn{3}{c}{\textbf{Hsb $\rightarrow$ Deu}} \\
            \cmidrule(lr){3-5} \cmidrule(lr){6-8}
        \multirow{4}*{60k} & \psp & 0.87 & 0.93 & 0.97 & 0.91 & 0.95 & 1.00 \\
         & \usp & 0.88 & 0.94 & 0.98 & 0.91 & 0.96 & 0.99 \\
         & \pspem & 0.99 & 0.99 & 1.01 & 0.99 & 0.99 & 1.00 \\
        \bottomrule
    \end{tabular}
\end{table}

\begin{table}[H]
    \caption{Begin-of-word tokens in the vocabulary.}
    \label{tab:start-word}
    \centering
    \footnotesize
    \setlength{\tabcolsep}{3pt}
    \begin{tabular}{ll ccc ccc}
        \toprule
        \multicolumn{8}{c}{\textbf{Start Word}} \\        
        \midrule
        Size & Model & 8k & 16k & 32k & 8k & 16k & 32k \\
        \midrule
        & & \multicolumn{3}{c}{\textbf{Fra $\rightarrow$ Ita}} & 
            \multicolumn{3}{c}{\textbf{Ita $\rightarrow$ Fra}} \\
            \cmidrule(lr){3-5} \cmidrule(lr){6-8}
        \multirow{4}*{100k} & \psp & 0.95 & 0.97 & 0.90 & 0.91 & 0.90 & 0.80 \\
         & \usp & 0.95 & 0.97 & 0.90 & 0.91 & 0.90 & 0.80 \\
         & \pspem & 0.40 & 0.37 & 0.35 & 0.38 & 0.36 & 0.33 \\
        & \sptgt & 0.80 & 0.81 & 0.75 & 0.79 & 0.80 & 0.72 \\
        \midrule
        \multirow{4}*{500k} & \psp & 0.93 & 0.96 & 0.96 & 0.92 & 0.94 & 0.92 \\
         & \usp & 0.93 & 0.96 & 0.96 & 0.92 & 0.94 & 0.92 \\
         & \pspem & 0.39 & 0.38 & 0.35 & 0.39 & 0.38 & 0.35 \\
        & \sptgt & 0.79 & 0.84 & 0.84 & 0.79 & 0.83 & 0.82 \\
        \midrule
        \multirow{4}*{1M} & \psp & 0.91 & 0.95 & 0.96 & 0.90 & 0.93 & 0.92 \\
         & \usp & 0.91 & 0.95 & 0.96 & 0.90 & 0.93 & 0.92 \\
         & \pspem & 0.39 & 0.37 & 0.36 & 0.38 & 0.37 & 0.34 \\
        & \sptgt & 0.78 & 0.84 & 0.85 & 0.76 & 0.82 & 0.84 \\
        \midrule
        & & \multicolumn{3}{c}{\textbf{Ces $\rightarrow$ Ukr}} & 
            \multicolumn{3}{c}{\textbf{Ukr $\rightarrow$ Ces}} \\
            \cmidrule(lr){3-5} \cmidrule(lr){6-8}
        \multirow{4}*{100k} & \psp & 0.95 & 0.97 & 0.98 & 0.94 & 0.94 & 0.95 \\
         & \usp & 0.95 & 0.97 & 0.98 & 0.94 & 0.94 & 0.95 \\
         & \pspem & 0.39 & 0.35 & 0.34 & 0.41 & 0.37 & 0.36 \\
        & \sptgt & 0.76 & 0.81 & 0.80 & 0.79 & 0.84 & 0.82 \\
        \midrule
        \multirow{4}*{500k} & \psp & 0.93 & 0.96 & 0.98 & 0.93 & 0.94 & 0.96 \\
         & \usp & 0.93 & 0.96 & 0.98 & 0.93 & 0.94 & 0.96 \\
         & \pspem & 0.40 & 0.37 & 0.36 & 0.42 & 0.38 & 0.38 \\
        & \sptgt & 0.73 & 0.80 & 0.84 & 0.76 & 0.83 & 0.87 \\
        \midrule
        \multirow{4}*{1M} & \psp & 0.91 & 0.95 & 0.97 & 0.91 & 0.94 & 0.95 \\
         & \usp & 0.91 & 0.95 & 0.97 & 0.91 & 0.94 & 0.95 \\
         & \pspem & 0.39 & 0.36 & 0.36 & 0.42 & 0.38 & 0.38 \\
        & \sptgt & 0.71 & 0.79 & 0.84 & 0.73 & 0.81 & 0.86 \\
        \midrule
        & & \multicolumn{3}{c}{\textbf{Ita $\rightarrow$ Mlt}} & 
            \multicolumn{3}{c}{\textbf{Mlt $\rightarrow$ Ita}} \\
            \cmidrule(lr){3-5} \cmidrule(lr){6-8}
        \multirow{4}*{100k} & \psp & 0.95 & 0.94 & 0.92 & 0.95 & 0.96 & 0.96 \\
         & \usp & 0.95 & 0.94 & 0.92 & 0.95 & 0.96 & 0.96 \\
         & \pspem & 0.39 & 0.35 & 0.35 & 0.40 & 0.36 & 0.36 \\
        & \sptgt & 0.73 & 0.75 & 0.73 & 0.81 & 0.83 & 0.80 \\
        \midrule
        & & \multicolumn{3}{c}{\textbf{Deu $\rightarrow$ Hsb}} & 
            \multicolumn{3}{c}{\textbf{Hsb $\rightarrow$ Deu}} \\
            \cmidrule(lr){3-5} \cmidrule(lr){6-8}
        \multirow{4}*{60k} & \psp & 0.98 & 0.98 & 0.99 & 0.96 & 0.95 & 0.84 \\
         & \usp & 0.98 & 0.98 & 0.99 & 0.96 & 0.95 & 0.84 \\
         & \pspem & 0.38 & 0.34 & 0.34 & 0.37 & 0.31 & 0.31 \\
        & \sptgt & 0.85 & 0.88 & 0.84 & 0.71 & 0.73 & 0.66 \\
        \bottomrule
    \end{tabular}
\end{table}

\begin{table}[t]
    \caption{Eflomal scores of the aligned text.}
    \label{tab:eflomal}
    \centering
    \footnotesize
    \setlength{\tabcolsep}{3pt}
    \begin{tabular}{ll ccc ccc}
        \toprule
        \multicolumn{8}{c}{\textbf{Eflomal scores ($\downarrow$)}} \\        
        \midrule
        Size & Model & 8k & 16k & 32k & 8k & 16k & 32k \\
        \midrule
        & & \multicolumn{3}{c}{\textbf{Fra $\rightarrow$ Ita}} & 
            \multicolumn{3}{c}{\textbf{Ita $\rightarrow$ Fra}} \\
            \cmidrule(lr){3-5} \cmidrule(lr){6-8}
        \multirow{2}*{1M} & \psp & 5.61 & 5.54 & 5.36 & 5.36 & 5.25 & 5.08 \\
        & \sptgt & 5.07 & 4.99 & 4.89 & 4.93 & 4.82 & 4.75 \\
        \midrule
        & & \multicolumn{3}{c}{\textbf{Ces $\rightarrow$ Ukr}} & 
            \multicolumn{3}{c}{\textbf{Ukr $\rightarrow$ Ces}} \\
            \cmidrule(lr){3-5} \cmidrule(lr){6-8}
        \multirow{2}*{1M} & \psp & 6.54 & 6.96 & 6.98 & 6.43 & 6.70 & 6.64 \\
        & \sptgt & 6.02 & 6.08 & 6.03 & 5.90 & 5.98 & 5.89 \\
        \midrule
        & & \multicolumn{3}{c}{\textbf{Ita $\rightarrow$ Mlt}} & 
            \multicolumn{3}{c}{\textbf{Mlt $\rightarrow$ Ita}} \\
            \cmidrule(lr){3-5} \cmidrule(lr){6-8}
        \multirow{2}*{100k} & \psp & 6.10 & 5.97 & 5.89 & 6.11 & 6.31 & 6.12 \\
        & \sptgt & 5.46 & 5.39 & 5.36 & 5.51 & 5.55 & 5.43 \\
        \midrule
        & & \multicolumn{3}{c}{\textbf{Deu $\rightarrow$ Hsb}} & 
            \multicolumn{3}{c}{\textbf{Hsb $\rightarrow$ Deu}} \\
            \cmidrule(lr){3-5} \cmidrule(lr){6-8}
        \multirow{2}*{60k} & \psp & 4.61 & 4.52 & 4.14 & 4.70 & 4.29 & 3.84 \\
        & \sptgt & 3.47 & 3.34 & 3.16 & 3.53 & 3.58 & 3.35 \\
        \bottomrule
    \end{tabular}
\end{table}

\twocolumn[\section{Machine Translation Evaluation}\label{app:mt}]
\begin{table}[H]
    \centering
    \footnotesize
    \setlength{\tabcolsep}{3pt}
    \caption{Average BLEU scores on the different language pairs and vocabulary sizes.}
    \label{tab:bleu}
    \begin{tabular}{l ccc ccc}
        \toprule
        \multicolumn{7}{c}{\textbf{BLEU ($\uparrow$)}} \\
        \midrule
        Model & 8k & 16k & 32k & 8k & 16k & 32k \\
        \midrule
        & \multicolumn{3}{c}{\textbf{Fra $\rightarrow$ Ita}} & \multicolumn{3}{c}{\textbf{Ita $\rightarrow$ Fra}} \\ 
            \cmidrule(lr){2-4} \cmidrule(lr){5-7}
        \spsrc + \psp & 24.5 & 24.2 & 22.9 &
            25.1 & 24.7 & 23.8 \\
        \spsrc + \sptgt & 25.1 & 24.8 & 23.1 &
            25.8 & 25.3 & 23.5 \\
        \midrule
        & \multicolumn{3}{c}{\textbf{Ces $\rightarrow$ Ukr}} &
            \multicolumn{3}{c}{\textbf{Ukr $\rightarrow$ Ces}} \\ \cmidrule(lr){2-4} \cmidrule(lr){5-7}
        \spsrc + \psp & 12.5 & 12.4 & 10.6 &
            18.9 & 19.1 & 19.2 \\
        \spsrc + \sptgt & 20.0 & 19.8 & 18.8 &
            21.6 & 21.2 & 19.6 \\
        \midrule
        & \multicolumn{3}{c}{\textbf{Ita $\rightarrow$ Mlt}} &
            \multicolumn{3}{c}{\textbf{Mlt $\rightarrow$ Ita}} \\ \cmidrule(lr){2-4} \cmidrule(lr){5-7}
        \spsrc + \psp & 0.3 & 5.9 & 5.6 &
            12.5 & 18.0 & 17.3 \\
        \spsrc + \sptgt & 6.0 & 5.9 & 5.4 &
            18.7 & 18.3 & 17.0 \\
        \midrule
        & \multicolumn{3}{c}{\textbf{Deu $\rightarrow$ Hsb}} &
            \multicolumn{3}{c}{\textbf{Hsb $\rightarrow$ Deu}} \\ \cmidrule(lr){2-4} \cmidrule(lr){5-7}
        \spsrc + \psp & 13.8 & 31.0 & 33.4 &
            12.1 & 37.2 & 37.4 \\
        \spsrc + \sptgt & 43.3 & 43.5 & 35.2 &
            41.4 & 41.9 & 29.4 \\
        \bottomrule
    \end{tabular}
\end{table}

\begin{table}[H]
    \centering
    \footnotesize
    \setlength{\tabcolsep}{3pt}
    \caption{Average TER scores on the different language pairs and vocabulary sizes.}
    \label{tab:ter)}
    \begin{tabular}{l ccc ccc}
        \toprule
        \multicolumn{7}{c}{\textbf{TER ($\downarrow$)}} \\
        \midrule
        Model & 8k & 16k & 32k & 8k & 16k & 32k \\
        \midrule
        & \multicolumn{3}{c}{\textbf{Fra $\rightarrow$ Ita}} & \multicolumn{3}{c}{\textbf{Ita $\rightarrow$ Fra}} \\ 
            \cmidrule(lr){2-4} \cmidrule(lr){5-7}
        \spsrc + \psp & 78.9 & 79.4 & 80.3 &
            82.5 & 83.2 & 84.4 \\
        \spsrc + \sptgt & 78.6 & 78.9 & 81.0 &
            82.2 & 82.6 & 85.4 \\
        \midrule
        & \multicolumn{3}{c}{\textbf{Ces $\rightarrow$ Ukr}} &
            \multicolumn{3}{c}{\textbf{Ukr $\rightarrow$ Ces}} \\ \cmidrule(lr){2-4} \cmidrule(lr){5-7}
        \spsrc + \psp & 105.4 & 115.6 & 119.6 &
            86.5 & 86.7 & 86.2 \\
        \spsrc + \sptgt & 89.7 & 90.2 & 92.4 &
            84.4 & 84.8 & 87.0 \\
        \midrule
        & \multicolumn{3}{c}{\textbf{Ita $\rightarrow$ Mlt}} &
            \multicolumn{3}{c}{\textbf{Mlt $\rightarrow$ Ita}} \\ \cmidrule(lr){2-4} \cmidrule(lr){5-7}
        \spsrc + \psp & 317.3 & 138.7 & 140.9 &
            137.8 & 84.9 & 86.1 \\
        \spsrc + \sptgt & 138.6 & 139.6 & 143.1 &
            84.4 & 84.9 & 86.9 \\
        \midrule
        & \multicolumn{3}{c}{\textbf{Deu $\rightarrow$ Hsb}} &
            \multicolumn{3}{c}{\textbf{Hsb $\rightarrow$ Deu}} \\ \cmidrule(lr){2-4} \cmidrule(lr){5-7}
        \spsrc + \psp & 85.0 & 71.6 & 71.8 &
            94.8 & 65.3 & 67.6 \\
        \spsrc + \sptgt & 59.7 & 60.0 & 71.6 &
            61.7 & 61.6 & 78.9 \\
        \bottomrule
    \end{tabular}

\end{table}

\begin{table}[H]
    \centering
    \footnotesize
    \setlength{\tabcolsep}{3pt}
    \caption{Average Comet scores on the different language pairs and vocabulary sizes. *: Maltese and Upper Sorbian are not included in the Comet training.}
    \label{tab:comet}
    \begin{tabular}{l ccc ccc}
        \toprule
        \multicolumn{7}{c}{\textbf{COMET ($\uparrow$)}} \\
        \midrule
        Model & 8k & 16k & 32k & 8k & 16k & 32k \\
        \midrule
        & \multicolumn{3}{c}{\textbf{Fra $\rightarrow$ Ita}} & \multicolumn{3}{c}{\textbf{Ita $\rightarrow$ Fra}} \\ 
            \cmidrule(lr){2-4} \cmidrule(lr){5-7}
        \spsrc + \psp & 0.797 & 0.799 & 0.786 &
            0.750 & 0.753 & 0.741 \\
        \spsrc + \sptgt & 0.801 & 0.805 & 0.780 &
            0.755 & 0.760 & 0.737 \\
        \midrule
        & \multicolumn{3}{c}{\textbf{Ces $\rightarrow$ Ukr}} &
            \multicolumn{3}{c}{\textbf{Ukr $\rightarrow$ Ces}} \\ \cmidrule(lr){2-4} \cmidrule(lr){5-7}
        \spsrc + \psp & 0.644 & 0.645 & 0.611 &
            0.711 & 0.714 & 0.726 \\
        \spsrc + \sptgt & 0.799 & 0.802 & 0.788 &
            0.756 & 0.757 & 0.734 \\
        \midrule
        & \multicolumn{3}{c}{\textbf{Ita $\rightarrow$ Mlt*}} &
            \multicolumn{3}{c}{\textbf{Mlt* $\rightarrow$ Ita}} \\ \cmidrule(lr){2-4} \cmidrule(lr){5-7}
        \spsrc + \psp & 0.436 & 0.590 & 0.591 &
            0.521 & 0.624 & 0.609 \\
        \spsrc + \sptgt & 0.592 & 0.592 & 0.592 &
            0.634 & 0.634 & 0.607 \\
        \midrule
        & \multicolumn{3}{c}{\textbf{Deu $\rightarrow$ Hsb*}} &
            \multicolumn{3}{c}{\textbf{Hsb* $\rightarrow$ Deu}} \\ \cmidrule(lr){2-4} \cmidrule(lr){5-7}
        \spsrc + \psp & 0.516 & 0.604 & 0.619 &
            0.409 & 0.606 & 0.602 \\
        \spsrc + \sptgt & 0.667 & 0.670 & 0.637 &
            0.641 & 0.651 & 0.546 \\
        \bottomrule
    \end{tabular}

\end{table}

\end{document}